\pdfoutput=1

\documentclass[11pt]{article}

\usepackage{EMNLP2023}

\usepackage{times}
\usepackage{latexsym}

\usepackage[T1]{fontenc}

\usepackage[utf8]{inputenc}

\usepackage{microtype}

\usepackage{inconsolata}

\usepackage{algorithm,algorithmic,xcolor}
\usepackage{algorithmic}

\usepackage{pifont}
\usepackage{amssymb,mathabx}
\usepackage{boldline} 
\usepackage{hyperref}
\usepackage{url}
\usepackage{enumitem}
\usepackage{array}
\usepackage{balance}
\usepackage{color}
\usepackage[toc,page]{appendix}
\usepackage{bbm}
\usepackage[noadjust,nosort]{cite}
\usepackage{booktabs}
\usepackage[normalem]{ulem}
\usepackage{setspace}
\usepackage{stfloats}
\usepackage{subcaption}
\usepackage{epsfig}
\usepackage{comment}
\usepackage{multirow}
\usepackage{wrapfig}
\usepackage{comment}

\title{ArchBERT: Bi-Modal Understanding of Neural Architectures and Natural Languages}

\author{Mohammad Akbari, Saeed Ranjbar Alvar, Behnam Kamranian, \\ {\bf Amin Banitalebi-Dehkordi, Yong Zhang} \\
         Huawei Technologies Canada Co., Ltd. \\
  \texttt{\{mohammad.akbari, saeed.ranjbar.alvar1, behnam.kamranian,} \\ \texttt{amin.banitalebi, yong.zhang3\}}@huawei.com}

\begin{document}
\maketitle
\begin{abstract}
Building multi-modal language models has been a trend in the recent years, where additional modalities such as image, video, speech, etc.  are jointly learned along with natural languages (i.e., textual information). Despite the success of these multi-modal language models with different modalities, there is no existing solution for neural network architectures and natural languages. Providing neural architectural information as a new modality allows us to provide fast architecture-2-text and text-2-architecture retrieval/generation services on the cloud with a single inference. Such solution is valuable in terms of helping beginner and intermediate ML users to come up with better neural architectures or AutoML approaches with a simple text query. In this paper, we propose ArchBERT, a bi-modal model for joint learning and understanding of neural architectures and natural languages, which opens up new avenues for research in this area. We also introduce a pre-training strategy named Masked Architecture Modeling (MAM) for a more generalized joint learning. Moreover, we introduce and publicly release two new bi-modal datasets for training and validating our methods. The ArchBERT's performance is verified through a set of numerical experiments on different downstream tasks such as architecture-oriented reasoning, question answering, and captioning (summarization). Datasets, codes, and demos are available \href{https://developer.huaweicloud.com/develop/aigallery/notebook/detail?id=e6a924c7-735a-4e02-a25b-4416b77b6315}{here}.
\end{abstract}

\section{Introduction}
\label{sec:introduction}
Existing machine learning models are mostly based on uni-modal learning, where a single modality is learned for the desired tasks. Example scenarios include image classification with image-only data; or language translation with text-only data \citep{t5, elang, gpt3}.
Despite the success of existing uni-modal learning methods at traditional single-modal tasks, they are usually insufficient \citep{baltruvsaitis2018multimodal} to model the complete aspects of human's reasoning and understanding of the environment. 

The alternative solution for this problem is to use multi-modal learning, where a model can jointly learn from multiple modalities such as text, image, or video to yield more abstract and generalized representations. As a result, a better understanding of various senses in information can be achieved and many new challenges that concern multi-modality can be handled. 
Such solution also enables the possibility of supplying a missing modality based on the observed ones. As an example, in text-based image generation, we aim to generate photo-realistic images which are semantically consistent with some given text description \citep{vlbeit}.

\begin{figure}
    \centering
    \includegraphics[width=1.0\columnwidth]{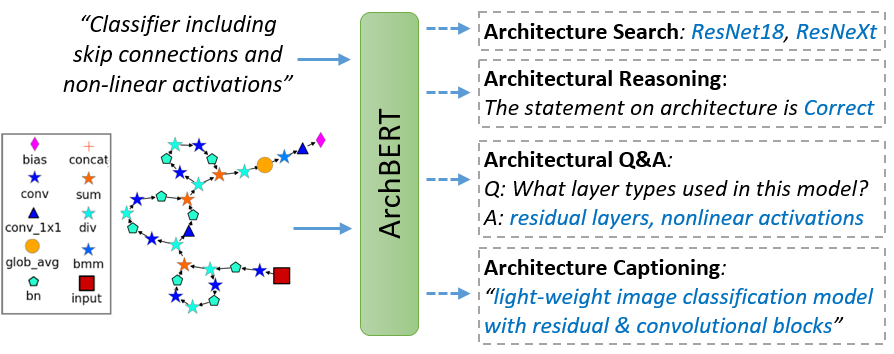}
    \vspace{-10pt}
    \caption{\label{fig:archbert-idea} Bi-modal understanding of neural architectures and natural languages with sample applications.
    }
    \vspace{10pt}
\end{figure}

One of the most popular multi-modal solutions is multi-modal language models (LMs), where an extra modality (e.g., image or video) is jointly used and learned along with the natural languages (i.e., textual information). Some of the recent multi-modal LMs include ViLBERT for image+text \citep{Vilbert}, VideoBERT for video+text \citep{videobert}, CodeBERT for code+text \citep{codebert}, and also GPT-4 \citep{openai2023gpt4}.

Although many multi-modal LMs with different modalities have been introduced so far, there is no existing solution for joint learning of neural network architectures and natural languages. Providing neural architectural information as a new modality allows us to perform many architecture-oriented tasks such as Architecture Search (AS), Architecture Reasoning (AR), Architectural Question Answering (AQA), and Architecture Captioning (AC) (Figure \ref{fig:archbert-idea}). 
{The real-world applications of such solution include fast architecture-2-text and text-2-architecture retrieval/generation services on the cloud with a single inference. Such solution is valuable in terms of helping users to come up with better neural architectures or AutoML approaches with a simple text query especially for beginner and intermediate ML users. For instance, AC can be used for automatically generating descriptions or model card information on a model hub (i.e., machine learning models repository). Furthermore, AR is helpful when a model is uploaded to a repository or cloud along with some textual description provided by the user, where the relevancy of the user's description for the given model can be automatically verified. If not verified, alternative auto-generated descriptions by a architecture-2-text solution can be proposed to the user.}

In this paper, we propose ArchBERT as a bi-modal solution for neural architecture and natural language understanding, where the semantics of both modalities and their relations can be jointly learned (Figure \ref{fig:archbert-idea}). 
To this end, we learn joint embeddings from the graph representations of architectures and their associated descriptions. Moreover, a pre-training strategy called Masked Architecture Modelling (MAM) for a more generalized and robust learning of architectures is proposed. We also introduce two new bi-modal datasets called TVHF and AutoNet for training and evaluating ArchBERT. To the best of our knowledge, ArchBERT is the first solution for joint learning of architecture-language modalities. In addition, ArchBERT can work with any natural languages and any type of neural network architectures designed for different machine learning tasks. The main contributions of this paper are as follows:
\vspace{-17pt}
\begin{itemize}[leftmargin=*]
    \setlength\itemsep{-1pt}
    \item A novel bi-modal model for joint learning of neural architectures and natural languages
    \item Two new bi-modal benchmark datasets for architecture-language learning and evaluation
    \item A new pre-training technique called MAM
    \item Introducing and benchmarking 6 architecture-language-related downstream applications
\end{itemize}


\begin{figure*}
    \centering
    \includegraphics[width=0.97\linewidth]{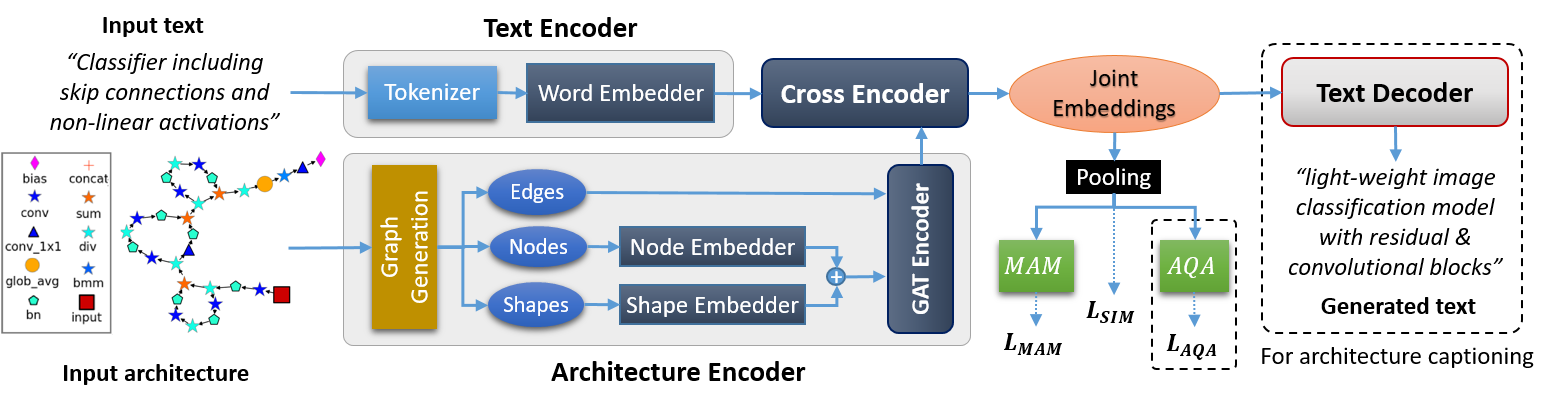}
    \vspace{-10pt}
    \caption{\label{fig:overall_framework} Overall framework of ArchBERT. 
    }
\end{figure*}

\vspace{-3pt}
\section{Related Works}
\label{sec:related-works}
Multi-modal models are used in many sub-fields in machine learning. For example, \citet{audiovisual1} and \citet{audiovisual3} introduced the audio-visual models trained on input acoustic speech signal and video frames of the speaker for speech enhancement, speech separation, and emotion recognition. Multi-modal models used in biomedical \citep{medical-multimodal1,medical-multimodal2}, remote-sensing \citep{remote-sense1,remote-sense2}, and autonomous driving \citep{AV1} applications have also proven to provide more accurate prediction and detection than the unimodal models. 

Among different types of multi-modal LMs in the literature, transformer-based ones have shown significant performance,
especially for vision-and-language tasks like visual question answering, image captioning, and visual reasoning. In VisualBERT \citep{visualbert}, a stack of transformers is used to align the elements of text and image pairs. ViLBERT \citep{Vilbert} extended BERT to a multi-modal double-stream model based on co-attentional transformer layers. 
In LXMERT \citep{LXMERT}, three encoders including language, object relation, and cross modality encoders are used. A single-stream vision-language model was introduced in VL-BEIT \citep{vlbeit}, where unpaired and paired image-text modalities were used for pre-training.

Video is another modality that is used with language in multi-modal models. VideoBERT \citep{videobert} is a single-stream video-language model, which learns a joint visual-linguistic representation from input video-text pairs. VIOLET \citep{violet} is another example that employs a video transformer to model the temporal dynamics of videos, and achieves SOTA results on video question answering and text-to-video retrieval. Programming language is also an emerging modality that has been used along with language. For example, CodeBERT \citep{codebert} is a multi-stream model, which uses LMs in each stream, where the input code is regarded as a sequence of tokens. On the other hand, GraphCodeBERT \citep{graphcodebert} proposes a structure-aware pre-training technique to consider the inherent structure of the code by mapping it to a data flow graph.

There are several prior works that combine more than two modalities. In Multimodal Transformer (MulT) \citep{unaligned-multimodal}, cross-modal attention modules are added to the transformers 
to learn representations from unaligned multi-modal streams, including the language, the facial gestures, and the acoustic behaviors. VATT \citep{vatt} also used video, audio, and text transformers along with a self-supervised learning strategy to obtain multi-modal representations from unlabeled data.

{It is worth mentioning that ChatGPT \citep{chatgpt} can be used for information retrieval, question answering, and also summarization over the textual descriptions of well-known neural architectures such AlexNet \citep{alexnet} or Faster-RCNN \citep{fasterrcnn}. However, unlike ArchBERT, it does not have a bi-modal understanding of both neural architectures (i.e., graphs) and natural languages, especially for newly proposed architectures and models.} 

\vspace{-3pt}
\section{Proposed Method: ArchBERT}
\label{sec:proposed-method}
The overall ArchBERT framework is shown in Figure \ref{fig:overall_framework}. The major components of ArchBERT include 
a text encoder, an architecture encoder, a cross encoder, and a pooling module. 

First, the input text represented by a sequence of $n$ words $W = \{w_i|i \in [1,n]\}$ is tokenized to a sequence of $n$ tokens $T = \{t_i|i \in [1,n]\}$.  
Then, the text encoder $E_t$ is utilized to map 
them to some word/token embeddings denoted by $M_t \in \mathbb{R}^{(n\times d)}$ with the embedding size of $d$:
$M_t = E_t(T).$

On the other hand, the architecture encoder is responsible for encoding the input neural architecture.
In this procedure, the computational graph of the input architecture is first extracted and represented with a directed acyclic graph $G=\{V,A,S\}$ where 
$V = \{v_i|i \in [1,m]\}$ denotes a sequence of $m$ nodes representing the operations and layers (e.g., convolutions, fully-connected layers, summations, etc.) and $A \in {\{0,1\}}^{m\times m}$ denotes a binary adjacency matrix describing the edges and the connectivity between the nodes. In addition to the nodes and edges, we also extract the shape of the parameters associated with each node (i.e., input/output channel dimensions and kernel sizes), denoted by $S = \{(s_i \in \mathbb{N}^{4})|i \in [1,m]\}$. 

The nodes and the shapes are separately encoded using the node and shape embedders $E_v$ and $E_s$, respectively. The adjacency matrix along with the summation 
of the resulting nodes and shapes embeddings are then given to a Graph Attention Network (GAT) \citep{gat} for computing the final architecture (graph) embeddings denoted by $M_g \in \mathbb{R}^{(m\times d)}$ with the embedding size of $d$:
\vspace{-4pt}
\begin{equation}
\label{eq:arch-encoder}
    M_g = GAT \big(E_v(V) + E_s(S), A\big)
    \vspace{-2pt}
\end{equation}

In general, GAT is designed to operate on graph-structured data in which a set of graph features (node+shape embeddings in our case) is transformed into higher-level features. Given the adjacency matrix, the GAT model also allows all nodes to attend over their neighborhoods’ features based on a self-attention strategy.

For joint learning of textual and architectural embeddings and share learning signals between both modalities, a cross transformer encoder, $E_c$, is used to process both embeddings in parallel.
These embeddings are then average-pooled to fixed-size 1D representations $J_t \in \mathbb{R}^{(1\times d)}$ and $J_g \in \mathbb{R}^{(1\times d)}$:
\vspace{-4pt}
\begin{equation}
\label{eq:cross-encoder}
    \{J_t, J_g\} = E_c (\{M_t, M_g\})
    \vspace{-2pt}
\end{equation}


As in S-BERT \citep{sbert}, we use the cosine similarity loss as a regression objective function to learn the similarity/dissimilarity between architectures and language embeddings. First, the cosine similarity between $J_t$ and $J_g$ are computed. Given a target soft score $y \in [0,1]$ (i.e., 0: dissimilar, 1: similar), the following mean squared-error (MSE) loss is then employed:
\vspace{-4pt}
\begin{equation}
    \label{eq:cos_loss}
    L_{SIM} = \|  y - \frac{J_t.J_g}{max(\|J_t\|_2.\|J_g\|_2, \epsilon)} \|_2,
    \vspace{-2pt}
\end{equation}
which minimizes the cosine distance between $J_t$ and $J_g$ pairs labeled as similar, while maximizes the distance for the dissimilar ones.

    





\subsection{Masked Architecture Modeling (MAM)} 
In the literature, a well-known pre-training objective function called Masked Language Modeling (MLM) is widely used by BERT-based models for learning language representations \citep{bert}. Inspired by MLM, we introduce a new objective called Masked Architecture Modeling (MAM) to provide more generalized learning and understanding of the graph embeddings corresponding to the neural architectures by ArchBERT. 



Inspired by BERT \citep{bert}, we randomly mask 15\% 
of the nodes with a special mask token and re-produce the masked nodes under the condition of the known ones. The MAM objective function is then defined as:
\vspace{-4pt}
\begin{equation}
\label{eq:mam-loss}
    L_{MAM} = -\mathbb{E}_{V_i \sim V}~log~ p(V_i | \hat{V}),
    \vspace{-2pt}
\end{equation}
where $\hat{V}$ is the masked version of $V$. In other words, $\hat{V}$ includes the contextual unmasked tokens surrounding the masked token $V_i$. In practice, the corresponding probability distribution is obtained by the MAM head $H_M$. The MAM head defines the distribution by performing the softmax function on the logits $F_m \in \mathbb{R}^{(m\times |\mathcal{E}|)}$ mapped from the graph embeddings $J_g$ as follows: 
$F_m = H_{M}(J_g),$
where $\mathcal{E}$ is the entire vocabulary of nodes (or nodes corpus) set. 
Given $L_{SIM}$ and $L_{MAM}$, the following weighted loss is then used for optimizing and pre-training the ArchBERT model:
\vspace{-4pt}
\begin{equation}
    \label{eq:total-loss}
    L = L_{SIM} + \alpha L_{MAM}.
    \vspace{-2pt}
\end{equation}



\subsection{Architectural Question Answering (AQA)}
\label{ssec:proposed-method-aqa}

The pre-trained ArchBERT can be utilized for the AQA task that is defined as the procedure of answering natural language questions about neural architectures. In other words, we can enable the ArchBERT model to predict the answers to architecture-related questions when the architecture and the question are matched.

For this task, we can fine-tune ArchBERT as a fusion encoder to jointly encode the input neural architecture and the question. To this end, the question and the architecture are first encoded using the text and architecture encoders, respectively. Both embeddings are then cross-encoded and pooled in order to calculate the final joint embeddings $J_t$ and $J_g$. 
The element-wise product is then computed to interactively catch similarity/dissimilarity and discrepancies between the embeddings. The resulting product is fed into AQA head for mapping to the logits $F_q \in \mathbb{R}^{|\mathcal{A}|}$ corresponding to $|\mathcal{A}|$ answers:
\vspace{-4pt}
\begin{equation}
\label{eq:aqa-head}
    F_q = H_{q}(J_t.J_g)
    \vspace{-2pt}
\end{equation}

As in \citep{anderson2018bottom}, the AQA in our work is formulated as a multi-label classification task, which assigns a soft target score to each answer based on its relevancy to $|\mathcal{A}|$ answers. A binary cross-entropy loss (denoted by $L_{AQA}$) on the target scores is then used as objective function.





\subsection{Language Decoder}
\label{ssec:proposed-method-ld}


We can empower the pre-trained ArchBERT to learn from and then benefiting for neural architecture captioning (or summarization) task by attaching a transformer decoder \citep{bart} to generate textual tokens one by one. 
In this regard, an auto-regressive decoding procedure is employed with the following loss function:
\vspace{-4pt}
\begin{equation}
\label{eq:langdec-loss}
    L_{DEC} = -\mathbb{E}_{T_i \sim T}~log~ p(T_i | T_{<i}, \hat{T}),
    \vspace{-3pt}
\end{equation}
where $\hat{T}$ is the masked version of the ground truth
text ${T}$, and $T_{i}$ is the $i$-th token to be predicted. $T_{<i}$ denotes the set of all the tokens decoded before $T_i$.
Similar to MAM, the probability distribution over the whole vocabulary is practically obtained by applying softmax on the decoded feature (or logits) $F_d \in \mathbb{R}^{(m\times |\mathcal{C}|)}$ that is calculated by providing the graph embeddings $J_g$ to the decoder:
$F_d = D_t(J_g)$, where $\mathcal{C}$ denotes the entire vocabulary set.



\vspace{-3pt}
\section{Datasets}
\vspace{-3pt}
\label{sec:datasets}
\begin{table*}[]
\caption{Statistical details of TVHF and AutoNet datasets (*: AQA, $\mu$: mean, $\sigma$: standard deviation, $M$: median).}
\vspace{-5pt}
\renewcommand{\arraystretch}{1.8}
\centering
\resizebox{\textwidth}{!}{%
\begin{tabular}{c|c|c|cccccccc|ccccccc}
\hlineB{3}
\multirow{3}{*}{\textbf{Dataset}} &
  \multirow{3}{*}{\textbf{Split}} &
  \multirow{3}{*}{\textbf{\#Samples}} &
  \multicolumn{8}{c|}{\textbf{Architecture}} &
  \multicolumn{7}{c}{\textbf{Text}} \\ \cline{4-18} 
 &
   &
   &
  \multicolumn{1}{c|}{\multirow{2}{*}{\textbf{\begin{tabular}[c]{@{}c@{}}\#Unique \\ Archs\end{tabular}}}} &
  \multicolumn{1}{c|}{\multirow{2}{*}{\textbf{\begin{tabular}[c]{@{}c@{}}\#Unique\\ Nodes\end{tabular}}}} &
  \multicolumn{3}{c|}{\textbf{\#Nodes}} &
  \multicolumn{3}{c|}{\textbf{\#Edges}} &
  \multicolumn{1}{c|}{\multirow{2}{*}{\textbf{\begin{tabular}[c]{@{}c@{}}\#Unique\\ Tokens\end{tabular}}}} &
  \multicolumn{3}{c|}{\textbf{\#Tokens}} &
  \multicolumn{3}{c}{\textbf{Sequence Length}} \\ \cline{6-11} \cline{13-18} 
 &
   &
   &
  \multicolumn{1}{c|}{} &
  \multicolumn{1}{c|}{} &
  \multicolumn{1}{c|}{\textbf{$\mu$}} &
  \multicolumn{1}{c|}{\textbf{$\sigma$}} &
  \multicolumn{1}{c|}{\textbf{$M$}} &
  \multicolumn{1}{c|}{\textbf{$\mu$}} &
  \multicolumn{1}{c|}{\textbf{$\sigma$}} &
  \textbf{$M$} &
  \multicolumn{1}{c|}{} &
  \multicolumn{1}{c|}{\textbf{$\mu$}} &
  \multicolumn{1}{c|}{\textbf{$\sigma$}} &
  \multicolumn{1}{c|}{\textbf{$M$}} &
  \multicolumn{1}{c|}{\textbf{$\mu$}} &
  \multicolumn{1}{c|}{\textbf{$\sigma$}} &
  \textbf{$M$} \\ \hline
\multirow{2}{*}{\textbf{TVHF}} &
  Train &
  24069 &
  \multicolumn{1}{c|}{538} &
  \multicolumn{1}{c|}{50} &
  \multicolumn{1}{c|}{1146.61} &
  \multicolumn{1}{c|}{1162.38} &
  \multicolumn{1}{c|}{705} &
  \multicolumn{1}{c|}{1281} &
  \multicolumn{1}{c|}{1302.90} &
  753 &
  \multicolumn{1}{c|}{3507} &
  \multicolumn{1}{c|}{16.16} &
  \multicolumn{1}{c|}{11.22} &
  \multicolumn{1}{c|}{14} &
  \multicolumn{1}{c|}{97.60} &
  \multicolumn{1}{c|}{77.76} &
  81 \\ \cline{2-18} 
 &
  Val &
  6018 &
  \multicolumn{1}{c|}{538} &
  \multicolumn{1}{c|}{50} &
  \multicolumn{1}{c|}{1146.61} &
  \multicolumn{1}{c|}{1162.38} &
  \multicolumn{1}{c|}{705} &
  \multicolumn{1}{c|}{1281} &
  \multicolumn{1}{c|}{1302.90} &
  753 &
  \multicolumn{1}{c|}{2965} &
  \multicolumn{1}{c|}{16.21} &
  \multicolumn{1}{c|}{11.59} &
  \multicolumn{1}{c|}{14} &
  \multicolumn{1}{c|}{97.88} &
  \multicolumn{1}{c|}{80.33} &
  81 \\ \hline
\multirow{2}{*}{\textbf{AutoNet}} &
  Train &
  103306 &
  \multicolumn{1}{c|}{10000} &
  \multicolumn{1}{c|}{28} &
  \multicolumn{1}{c|}{371.50} &
  \multicolumn{1}{c|}{312.61} &
  \multicolumn{1}{c|}{266} &
  \multicolumn{1}{c|}{401} &
  \multicolumn{1}{c|}{322.99} &
  241 &
  \multicolumn{1}{c|}{769} &
  \multicolumn{1}{c|}{43.81} &
  \multicolumn{1}{c|}{8.62} &
  \multicolumn{1}{c|}{45} &
  \multicolumn{1}{c|}{333.67} &
  \multicolumn{1}{c|}{74.80} &
  345 \\ \cline{2-18} 
 &
  Val &
  10338 &
  \multicolumn{1}{c|}{1000} &
  \multicolumn{1}{c|}{28} &
  \multicolumn{1}{c|}{384.48} &
  \multicolumn{1}{c|}{343.31} &
  \multicolumn{1}{c|}{266} &
  \multicolumn{1}{c|}{419} &
  \multicolumn{1}{c|}{368.20} &
  293.5 &
  \multicolumn{1}{c|}{652} &
  \multicolumn{1}{c|}{43.92} &
  \multicolumn{1}{c|}{8.66} &
  \multicolumn{1}{c|}{45} &
  \multicolumn{1}{c|}{334.01} &
  \multicolumn{1}{c|}{74.92} &
  345 \\ \hline
\multirow{2}{*}{\textbf{AutoNet*}} &
  Train &
  350000 &
  \multicolumn{1}{c|}{10000} &
  \multicolumn{1}{c|}{28} &
  \multicolumn{1}{c|}{373.33} &
  \multicolumn{1}{c|}{313.90} &
  \multicolumn{1}{c|}{270} &
  \multicolumn{1}{c|}{404} &
  \multicolumn{1}{c|}{325.45} &
  297 &
  \multicolumn{1}{c|}{86} &
  \multicolumn{1}{c|}{10.78} &
  \multicolumn{1}{c|}{1.89} &
  \multicolumn{1}{c|}{11} &
  \multicolumn{1}{c|}{62.76} &
  \multicolumn{1}{c|}{12.48} &
  62 \\ \cline{2-18} 
 &
  Val &
  35000 &
  \multicolumn{1}{c|}{1000} &
  \multicolumn{1}{c|}{28} &
  \multicolumn{1}{c|}{358.3} &
  \multicolumn{1}{c|}{301.98} &
  \multicolumn{1}{c|}{261} &
  \multicolumn{1}{c|}{390} &
  \multicolumn{1}{c|}{324.31} &
  285.5 &
  \multicolumn{1}{c|}{86} &
  \multicolumn{1}{c|}{10.79} &
  \multicolumn{1}{c|}{1.89} &
  \multicolumn{1}{c|}{11} &
  \multicolumn{1}{c|}{62.76} &
  \multicolumn{1}{c|}{12.45} &
  62 \\ \hlineB{3}
\end{tabular}%
}
\label{tbl:datasets-stats}
\end{table*}

For pre-training the ArchBERT model, a dataset of neural architectures labeled with some relevant descriptions is required. To the best of our knowledge, there is no such bi-modal dataset in the literature. In this paper, we introduce two datasets called TVHF and AutoNet for bi-modal learning of neural architectures and natural languages.
The numerical and the statistical details of TVHF and AutoNet datasets are summarized in Table \ref{tbl:datasets-stats}.

{
Note that all the labels and descriptions in the proposed datasets have been manually checked and refined by human. There may be some minor noise in the dataset (i.e., an inevitable nature of any dataset, especially the very first versions), but in overall, the datasets are of sufficient quality for our proof-of-concept experiments.
}

\vspace{-3pt}
\subsection{TVHF}
In order to create this dataset, we collected 538 unique neural architectures form TorchVision (TV) \citep{torchvision} and HuggingFace (HF) \citep{huggingface} frameworks. The descriptions relevant to the architectures were extracted from TV and HF frameworks as well as other online resources such as papers and web pages (with the vocabulary size $|\mathcal{C}|$=31,764). To increase the dataset size, the descriptions were split into individual sentences each assigned to the related architecture, which provided a collection of 2,224 positive samples, i.e., pairs of architecture with their relevant descriptions (details in the appendix). 

To assure the model learns both similarities and dissimilarities, we also generated negative samples by assigning irrelevant descriptions to the architectures (resulting in a total of 27,863 negative samples). We randomly split the dataset (in total 30,087 samples) into 80\% for train and 20\% for validation. 

For fine-tuning and evaluating ArchBERT on Architecture Clone Detection (ACD), we establish another dataset including pairs of architectures manually hard-labeled with a dissimilarity/similarity score (0 or 1). To this end, 
all combinations of two architectures from TVHF were collected (in total 82.8K samples) and split into train/val sets (80\% and 20\%). Details are provided in the appendix.



\subsection{AutoNet}



As described before, TVHF includes realistic human-designed architectures, which are manually labeled with real descriptions. On the other hand, we introduce the AutoNet dataset, which includes automatically generated architectures and descriptions. AutoNet is basically 
the modified and extended version of DeepNet1M \citep{ppuda}, which is a standardized benchmark and dataset of randomly generated architectures for the parameter prediction tasks.

\begin{figure}
    \centering
    \includegraphics[width=\linewidth]{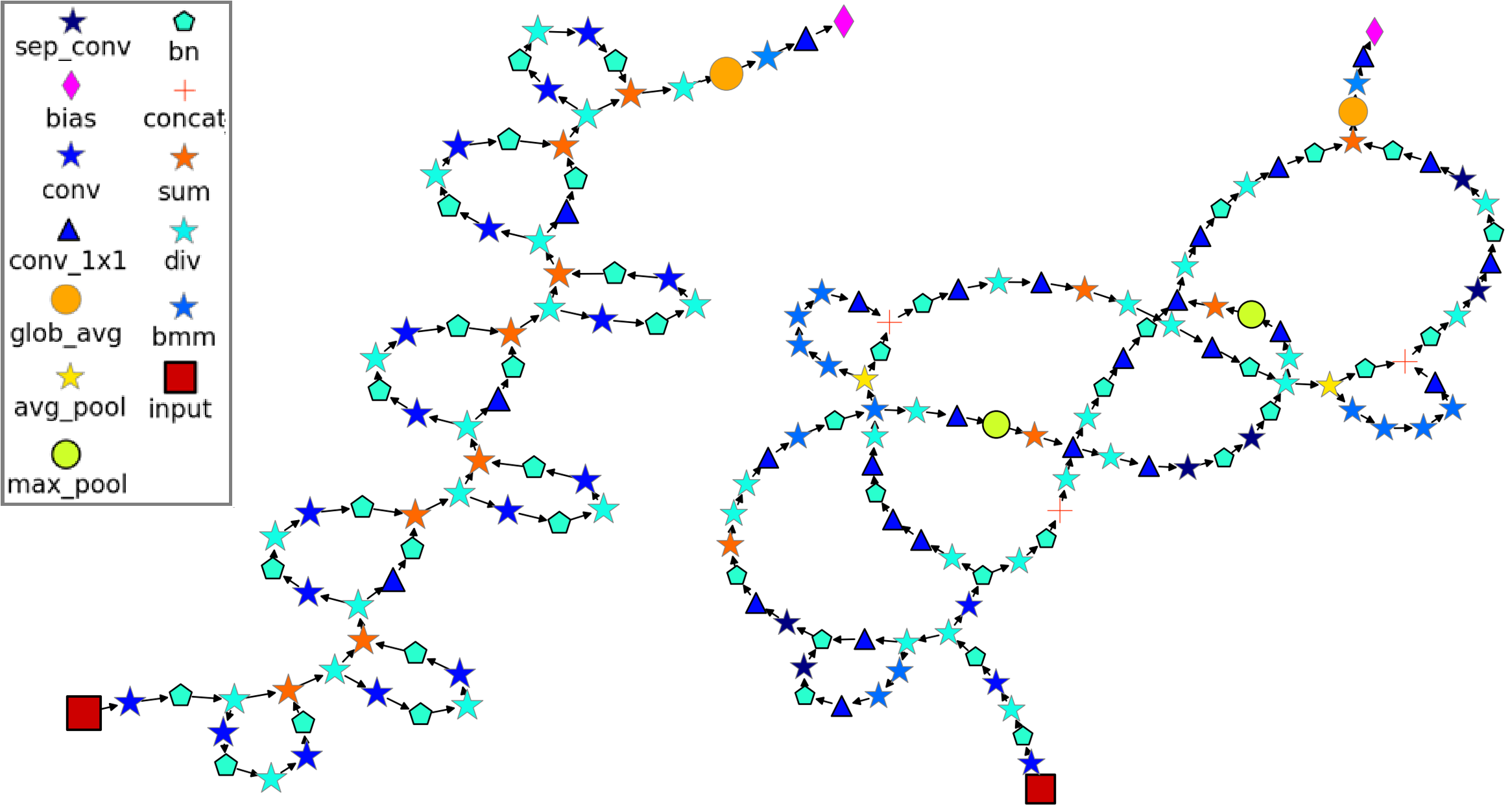}
    \caption{\label{fig:sample-graphs} 
    Sample graphs generated for ResNet18 (left) and a random architecture from AutoNet (right).
    }
    \vspace{5pt}
\end{figure}

In AutoNet, we extend the set of operations (layers) from 15 types (in DeepNet1M) to 85, which include most of the recent operations used in computer vision and natural language models. We followed the same procedure in DeepNet1M and randomly generated 10K and 1K architectures for train and validation sets, respectively. 

For automatic generation of textual descriptions related to each architecture, we created an extensive set of sentence templates, which were filled based on the information extracted from the structure, modules, and existing layers of the corresponding architecture. The same process was applied for generating negative samples, but with the textual information of the non-existing modules and layers in the architecture. For each architecture, 10-11 textual descriptions were created, which resulted in 103,306 and 10,338 architecture and text pairs for the train and validation sets (with the vocabulary size $|\mathcal{C}|$=30,980), respectively. The details of this procedure are given in the appendix.

\vspace{-3pt}
\subsubsection{AutoNet-AQA}
For fine-tuning and evaluating ArchBERT on AQA, another dataset including triplets of architectures, questions, and answers is needed. 
As in AutoNet, a set of question/answer templates were used to automatically generate the questions and answers. The same procedure of generating neural architectures as in AutoNet was employed.
10K and 1K architectures were respectively created for the train and validation sets. For each architecture, 35 unique questions were generated, and the answers were chosen from a list of $|\mathcal{A}|=51$ unique answers. In total, the train and validation sets respectively include 350K and 35K samples. 

The visualization of two sample graphs generated for ResNet18 from TVHF and a random architecture from AutoNet is shown in Figure \ref{fig:sample-graphs}. {More sample data along with the quality analysis of the datasets are given in the appendix.
}

\vspace{-3pt}
\section{Experimental Results}
\label{sec:experimental-results}
In this section, the performance of ArchBERT on the following downstream tasks is evaluated and numerically analyzed.
\vspace{-7pt}
\begin{itemize}[leftmargin=*]
    \setlength\itemsep{-3pt}
    \item \textbf{Architectural Reasoning (AR)}: it is the task of determining if a statement regarding an architecture is correct or not. 
    \item \textbf{Architecture Clone Detection (ACD)}: it includes the process of checking if two architectures are semantically/structurally similar or not. 
    \item \textbf{Architectural Question Answering (AQA)}: as given in Section \ref{sec:proposed-method}, it is the process of providing an answer to a question over a given architecture.
    \item \textbf{Architecture Captioning (AC)}: it is the task of generating descriptions for a given architecture.
\end{itemize}

Since there is no related prior works, we compare our method with some uni-modal baselines for each of the above tasks. 
An ablation study over different components of ArchBERT is also presented.

In this work, we employ the BERT-Base model (with 12 heads) as our ArchBERT's cross encoder. We pre-trained ArchBERT on both TVHF and AutoNet datasets with a batch size of 80, embedding size of $d$=768, and the Adam optimizer with learning rate of 2e-5 for 6 hours.
The training on TVHF and AutoNet was respectively done for 20 and 10 epochs. Since there is a large scale difference between the $L_{SIM}$ and $L_{MAM}$ loss values in the weighted loss in Equation \ref{eq:total-loss}, where $L_{MAM}$$\gg$$L_{SIM}$, we set $\alpha$=5e-2 to balance the total loss value {(obtained experimentally)}. A batch size of 80 is used for all the tests with the pre-trained ArchBERT. 

{
}

\begin{figure*}[!tb]
    \centering
    \includegraphics[width=0.99\linewidth]{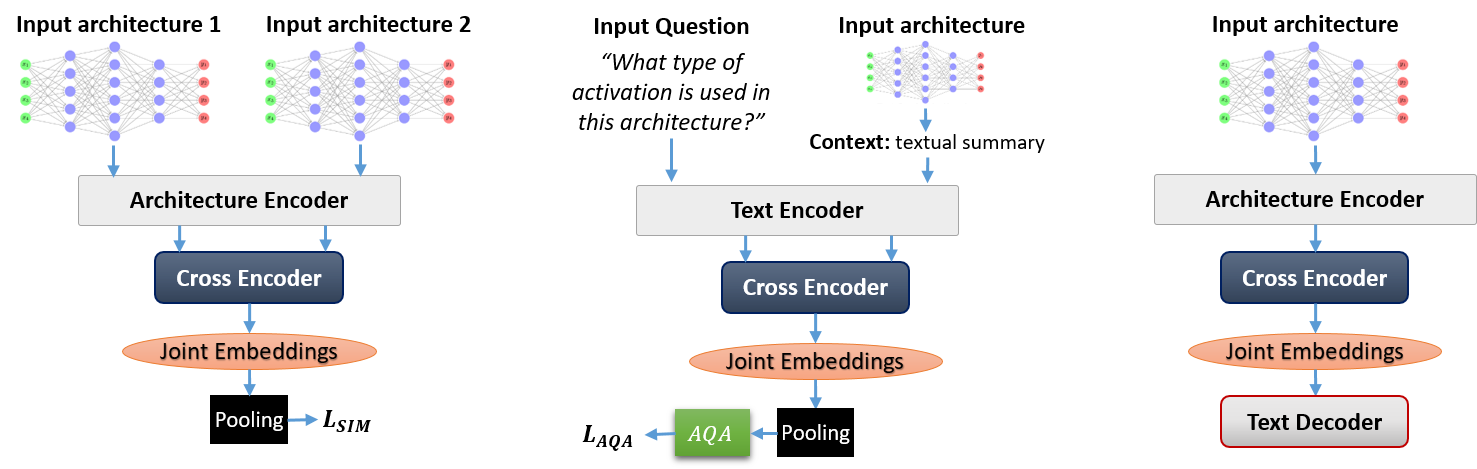}
    \caption{Uni-Modal Baselines (left: ACD, middle: AQA, right: AC).}
    \label{supp:baselines}
\end{figure*}

\subsection{Uni-Modal Baselines}
For the AR baseline, we compare the architecture name with an input statement, which is considered as "correct" if the architecture name appears in the statement, otherwise it is "incorrect". Note that unlike this baseline, ArchBERT does not need the architecture name to infer about the statements. 

For the ACD uni-modal baseline (Figure \ref{supp:baselines}-left), the architecture encoder is first used to separately map both input architectures, denoted by $\{G^1, G^2\}$, into the graph embeddings $\{M_g^1, M_g^2\}$ (Equation \ref{eq:arch-encoder}). The cross encoder and pooling module are then applied to obtain the fixed-size joint representations $\{J_g^1, J_g^2\}$ (Equation \ref{eq:cross-encoder}). The cosine similarity loss in Equation \ref{eq:cos_loss} is finally performed on $\{J_g^1, J_g^2\}$ pairs along with a provided hard-label. {For this baseline, we trained ArchBERT with architecture-only pairs (without text encoder) from TVHF-ACD train set.}

For the AQA uni-modal baseline (Figure \ref{supp:baselines}-middle), we train a text-only ArchBERT (without architecture encoder), where the context is obtained from the textual information and summary of the input architecture, e.g., layer names (i.e., using Pytorch model summary function). The extracted information is considered as the input context on which the question answering procedure is performed. The tokenized input question and context, denoted by $\{T^q, T^c\}$, are mapped into token embeddings $\{M_t^q, M_t^c\}$, which are then cross-encoded and average-pooled to obtain the joint embeddings $\{J_t^q, J_t^c\}$ (Equation \ref{eq:cross-encoder}). As in Equation \ref{eq:aqa-head}, the element-wise product of $\{J_t^q, J_t^c\}$ is given to the AQA head to obtain the logits required for the binary cross-entropy loss described in Section \ref{ssec:proposed-method-aqa}.

For the AC uni-modal baseline (Figure \ref{supp:baselines}-right), we trained ArchBERT (without text encoder) followed by the decoder from scratch (no bi-modal pre-training of ArchBERT). The detailed AC procedure is described in Section \ref{ssec:proposed-method-ld}.

\vspace{-3pt}
\subsection{Architectural Reasoning (AR)}
\label{ssec:experimental-results-ar}
For this task, the input text and the architecture are given to ArchBERT to create the pooled embeddings. The cosine similarity score between these embeddings is then computed. If the score is greater than some threshold $\tau$ 
(i.e., 0.5), the statement on the architecture is determined as “correct”, otherwise “incorrect”. We evaluate the performance of the pre-trained ArchBERT on this task over the TVHF validation set. As summarized in Table \ref{tbl:results}, an accuracy and F1 score of 96.13\% and 71.86\% were respectively achieved. F1 scores are reported to deal with the class imbalance.

As reported in Table \ref{tbl:results}, a F1 score of 55.93\% is achieved by the AR baseline, which is about 16\% lower than ArchBERT.

\vspace{-3pt}
\subsection{Architecture Clone Detection (ACD)}
\label{ssec:experimental-results-acd}
To perform this task, both input architectures are given to ArchBERT's architecture encoder followed by the cross-encoder and pooling module to obtain the pooled embeddings. The cosine similarity of the embeddings is then computed. If the similarity score is greater than a threshold (i.e., 0.5), the two architectures are considered similar, otherwise dissimilar. 

{
We first evaluate the pre-trained ArchBERT's performance} on the TVHF-ACD validation set. Although the pre-trained model has not specifically learned to detect similar/dissimilar architectures, it still achieves a good accuracy of 86.20\% and F1 score 60.10\% (Table \ref{tbl:results}). {However, by fine-tuning the pre-trained ArchBERT with TVHF-ACD train set, significantly improved accuracy and F1 score of 96.78\% and 85.98\% are achieved.}

{
Two baselines including Jaccard similarity \citep{jaccard2} and a uni-modal version of ArchBERT are used to compare with our bi-modal ArchBERT on ACD task.} For Jaccard, the similarity of the architecture pairs is computed by taking the average ratio of intersection over union of the nodes and edges ($V$ and $A$). The pairs are considered as "similar" if the similarity score is greater than 0.5, otherwise “dissimilar". As shown in Table \ref{tbl:results}, the pre-trained and fine-tuned ArchBERT models respectively outperform this baseline with 14\% and 40\% higher F1 scores. {
The ACD uni-modal baseline also achieves F1 score of 84\%, i.e., 2\% lower than fine-tuned ArchBERT.
}


\subsection{Architectural Question Answering (AQA)}
\label{ssec:experimental-results-aqa}
For this, ArchBERT along with the attached AQA head (composed of a two layer MLP) is fine-tuned with the AutoNet-AQA dataset using a batch size of 140 over 10 epochs (for about 10 hours). We use the Adam optimizer with an initial learning rate of 2e-5. At the inference time, we simply take a sigmoid over the AQA head's logits (with the same batch size of 140). As given in Table \ref{tbl:results}, ArchBERT achieves an accuracy of 72.73\% and F1 score of and 73.51\% over the AutoNet-AQA validation set.

\begin{table}[!tb]
\caption{\label{tbl:results} The performance of ArchBERT and its components on different tasks and datasets (AR: Architectural Reasoning, ACD: Architecture Clone Detection, AQA: Architectural Question Answering, CR: Cross Encoder, MAM: Masked Architecture Modeling).}
\vspace{-15pt}
\fontsize{9.5}{10}\selectfont
\begin{center}
\begin{tabular}[t]{p{0.6cm}p{1.0cm}p{2.4cm}p{0.8cm}p{0.8cm}}
\toprule
\textbf{Task} & \textbf{Dataset} & \textbf{Model} & \textbf{Acc(\%)} & \textbf{F1(\%)} 
\\
\\[-0.35cm]
\midrule
\hspace{-5pt} \multirow{6}{*}{AR} &\multirow{6}{*}{TVHF} & \textbf{ArchBERT} & \textbf{96.13} & \textbf{71.86} \\
\hspace{-5pt} & & {-w/o  {Shape}} & 95.44 & 69.16 \\
\hspace{-5pt} & & {-w/o  {Edge}} & 95.52 & 68.98 \\
\hspace{-5pt} & & {-w/o  {Edge+Shape}} & 95.12 & 65.80 \\
\hspace{-5pt} & & -w/o  {MAM} & 95.18	& 64.27 \\
\hspace{-5pt} & & -w/o  {CR} & 94.42 & 57.03\\
\hspace{-5pt} & &  {Baseline} & 89.03 & 55.93  \\
\midrule
\hspace{-5pt}  \multirow{9}{*}{ACD} &\multirow{9}{*}{TVHF} & \textbf{ArchBERT} & \textbf{86.20}	&\textbf{60.10}\\
\hspace{-5pt} & & {-w/o  {Shape}} & 85.44 & 60.20 \\
\hspace{-5pt} & & {-w/o  {Edge}} & 76.70 & 47.96 \\
\hspace{-5pt} & & {-w/o  {Edge+Shape}} & 82.90 & 56.45 \\
\hspace{-5pt} & & -w/o  {MAM}& 78.80 & 49.59 \\
\hspace{-5pt} & &-w/o  {CR} & 69.89 & 42.35	 \\
\hspace{-5pt} & &  {Jaccard} & 80.22 & 45.96	 \\
\hspace{-5pt} & &  \textbf{ArchBERT-ft} & \textbf{96.78} & {85.98}	 \\
\hspace{-5pt} & &  {Baseline (uni)} & 96.24 & 84.01\\
\midrule
\hspace{-5pt}  \multirow{4}{*}{AQA} & \multirow{4}{*}{AutoNet} & 
\textbf{ArchBERT} & \textbf{72.73}	&\textbf{73.51} \\
\hspace{-5pt} & &-w/o  {MAM}& 66.08 & 66.16 \\
\hspace{-5pt} & & -w/o  {CR}& 60.32 & 63.33 \\
\hspace{-5pt} & &  {Baseline (uni)} & 55.82 & 61.84	 \\
\bottomrule
\end{tabular}
\end{center}
\end{table}

For the AQA baseline, an F1 score of 61.84\% was obtained on AutoNet-AQA, which is $\approx$12\% lower than the proposed bi-modal ArchBERT.


\subsection{Architecture Captioning (AC)}
\label{ssec:experimental-results-ac}
To analyze ArchBERT's performance on AC, the pre-trained ArchBERT (without text encoder) attached with a language decoder is fine-tuned on both TVHF and AutoNet with a batch size of 30 for 10 epochs. The fine-tuning process for TVHF and AutoNet respectively took about 0.5 and 6 hours. Adam optimizer with an initial learning rate of 2e-5 was used. For the language decoder, a single-layer transformer decoder (with 12 heads and hidden size of $d$=768) followed by 2 linear layers is used.

At the inference, the beam search (with the size of 10) was employed to auto-regressively generate the output tokens, which were then decoded back to their corresponding words. The same batch size of 30 was used for the evaluation. The results over the TVHF and AutoNet validation sets are summarized in Table \ref{tbl:results-ld}, where Rouge-Lsum-Fmeasure (RL) \citep{lin2004rouge} scores of 0.17 and 0.46 were respectively achieved. Unlike AutoNet, TVHF dataset includes more complicated neural architectures along with high-level human-written textual descriptions, which makes the architecture captioning more challenging. As a result, lower performance is achieved. 

\begin{table}[!tb]
\caption{\label{tbl:results-ld} ArchBERT's performance on Architecture Captioning (AC) (CR: Cross Encoder, MAM: Masked Architecture Modeling, R1: Rouge1-Fmeasure, R2: Rouge2-Fmeasure, RL: Rouge-Lsum-Fmeasure).}
\vspace{-15pt}
\fontsize{9.5}{10}\selectfont
\begin{center}
\begin{tabular}[t]{p{1.1cm}p{2.0cm}p{0.85cm}p{0.85cm}p{0.85cm}}
\toprule
\textbf{Dataset}& \textbf{Model} & \textbf{R1} & \textbf{R2} & \textbf{RL}
\\
\\[-0.35cm]
\midrule
\hspace{-5pt}  \multirow{3}{*}{TVHF} & \textbf{ArchBERT} & \textbf{0.18}&	0.05	&\textbf{0.17} \\
\hspace{-5pt} & -w/o  {MAM} & 0.17	& {0.05} &	0.15\\
\hspace{-5pt} & {Baseline (uni)} & 0.18	& 0.07 & 0.17\\
\midrule
\hspace{-5pt}  \multirow{3}{*}{AutoNet\textbf{}} & \textbf{ArchBERT} & \textbf{0.48}&	\textbf{0.36}&	\textbf{0.46}
 \\
\hspace{-5pt} &-w/o  {MAM} & 0.45	&0.34&	0.43\\
\hspace{-5pt} & {Baseline (uni)} & 0.40	& 0.30 &	0.38\\
\bottomrule
\end{tabular}
\end{center}
\end{table}

{
The uni-modal AC baseline achieves an RL of 0.38 on AutoNet, which is 8\% lower than the proposed bi-modal ArchBERT (i.e., pre-trained on both architectures and text, and fine-tuned for AC).
}

\subsection{Architecture Search (AS)}
ArchBERT is also applicable to Architecture Search (AS) downstream task. The task is to design a semantic search engine to receive a textual query from the user, search over a database of numerous neural architectures (or models), and return the best matching ones. As for any semantic search engine, an indexed database of all searched architecture embeddings is needed, within which the architecture search is performed. For the search procedure over such database using ArchBERT, the text query is encoded by the text encoder, and then is cross-encoded to make sure the previously-learned architectural knowledge is also utilized for computing final text embeddings. The pooled text embeddings are then compared with all the architecture embeddings stored in the database to find the best matching (most similar) architectures. We did not report any numerical analysis for AS due to the lack of related validation set. However, qualitative demo is available in the supplementary materials.

\subsection{Qualitative Results}
In Table \ref{tbl:sample-results}, ArchBERT's predictions on AR and ACD tasks over some samples from TVHF validation set are given. In addition, we present the predictions on AC and AQA tasks over the right architecture in Figure \ref{fig:sample-graphs} (i.e., a sample from AutoNet validation set).
Sample cases for which ArchBERT makes wrong predictions are also given in the table (marked with *), e.g., AR's prediction for Vit\_b\_16 and ConvNext-tiny architectures. 


\begin{table}[tb!]
\footnotesize
\caption{\label{tbl:sample-results} Qualitative results on various tasks (\checkmark: Correct/Similar, \ding{55}: Incorrect/Dissimilar, *: wrong preds).}
\centering
\renewcommand{\arraystretch}{1.05}
{
\begin{tabular}{|l|l|c|c|}
\hline
\multicolumn{1}{|c|}{\textbf{Architecture}} & \multicolumn{1}{c|}{\textbf{Text}} & \textbf{AR} & \textbf{ACD} 
\\ 
\hline
ResNet18 & 
\begin{tabular}[c]{@{}l@{}}image classifier with\\ residual layers\end{tabular} & \checkmark &
\\ \cline{1-3}
\begin{tabular}[c]{@{}l@{}}Fasterrcnn\\(ResNet50)\end{tabular}
& \begin{tabular}[c]{@{}l@{}}text classifier using\\ bert-based models\end{tabular} & \ding{55} 
& \multirow{-2}{*}{\ding{55}} 
\\ 
\hline
\hline
{Bert-base} & \begin{tabular}[c]{@{}l@{}}object detection\\ for photos\end{tabular} & \ding{55} &
\\ \cline{1-3}
\begin{tabular}[c]{@{}l@{}}RoBERT\\(small)\end{tabular}
& \begin{tabular}[c]{@{}l@{}}text classifier using\\ bert-based models\end{tabular} & \checkmark
& \multirow{-2}{*}{\checkmark} 
\\ 
\hline
\hline
{Vit\_b\_16} & \begin{tabular}[c]{@{}l@{}}bert-like image\\ classification\end{tabular} & \ding{55}$^*$ &
\\ \cline{1-3}
\begin{tabular}[c]{@{}l@{}}Fasterrcnn\\(mobilenet)\end{tabular}
& 
\begin{tabular}[c]{@{}l@{}}object detection for\\ photos\end{tabular}
 & \checkmark
& \multirow{-3}{*}{\ding{55}} 
\\ 
\hline
\hline
\begin{tabular}[c]{@{}l@{}}ConvNext\\(tiny)\end{tabular} & \begin{tabular}[c]{@{}l@{}}a very large convnext\\ architecture\end{tabular} & \checkmark$^*$ &
\\ 
\cline{1-3}
{Bert-mini} & \begin{tabular}[c]{@{}l@{}}language model with\\ attention layers\end{tabular} & \checkmark 
& \multirow{-3}{*}{\ding{55}} 
\\

\hline
\hline
\hline
\multicolumn{1}{|l|}{\begin{tabular}[l]{@{}l@{}}Figure \ref{fig:sample-graphs}'s right\\ architecture\\ (AutoNet)\end{tabular}} & 
\multicolumn{3}{l|}{\begin{tabular}[l]{@{}l@{}}
\textbf{AC:}
"\textit{this model separable convolution}\\ \textit{which divides a single convolution into}\\ \textit{two convolutions}"\\
\hline
\textbf{AQA:} What type of pooling is used\\ in this architecture?\\ \textbf{Prediction}: \textit{'MaxPool2d', 'AvgPool2d'}
\end{tabular}} 
\\
\hline
\end{tabular}
}
\end{table}

\vspace{-3pt}
\subsection{Ablation Study}
\label{ssec:experimental-results-ablations}
We conduct ablation study to analyze the effect of ArchBERT's different modules such as MAM, Cross Encoder, and graph elements on the performance of AR, ACD, AQA, and AC tasks. The results are summarized in Tables \ref{tbl:results} and \ref{tbl:results-ld}.

First, we remove the MAM head and its loss from the pre-training and fine-tuning stages. The performance of the pre-trained model without MAM is evaluated on AR and ACD with the TVHF dataset. As seen in Table \ref{tbl:results}, excluding MAM in pre-training results in a significant 
F1 drops by 7.59\% and 10.51\% on AR and ACD tasks, respectively. The effect of MAM on finetuend ArchBERT for AQA and AC downstream tasks is also evaluated and reported in in Tables \ref{tbl:results} and \ref{tbl:results-ld}. It is shown that using MAM provides F1 score improvements of 7.35\% and 0.03\% on AQA and AC, respectively. 

We also study the ArchBERT's performance when the Transformer cross encoder is not used for encoding the architectures. In this case, the embeddings obtained from the architecture encoder are directly used for training and evaluating the model by bypassing the cross encoder. The corresponding results on AR, ACD, and AQA tasks are given in Table \ref{tbl:results}. From the results, when the cross encoder is removed, the performance of both the pre-trained and fine-tuned models decreases. This reveals the importance of the cross encoder in joint encoding and learning of the text and architecture. As seen in the table, the F1 scores on AR, ACD, and AQA tasks are substantially reduced by 14.83\%, 17.75\%, and 10.18\%, respectively, if the cross encoder is not utilized for architecture encoding.

We also ran a set of ablations over different graph items. For AR, F1 scores of 71.86\% (ArchBERT), 69.16\% (w/o shape), 68.98\% (w/o edge), and 65.80\% (w/o shape+edge) are achieved. For ACD, F1 scores of 60.10\% (ArchBERT), 60.20\% (w/o shape), 47.96\% (w/o edge), and 56.45\% (w/o shape+edge) are obtained. It is seen that using all graph items provides the best results. For ACD, the shape has no effect on F1 score, but excluding it gives $\approx$1\% lower accuracy.

{The ArchBERT's performance on out-of-distribution data will be presented in the appendix.}

\begin{figure*}
    \centering
    \includegraphics[width=\linewidth]{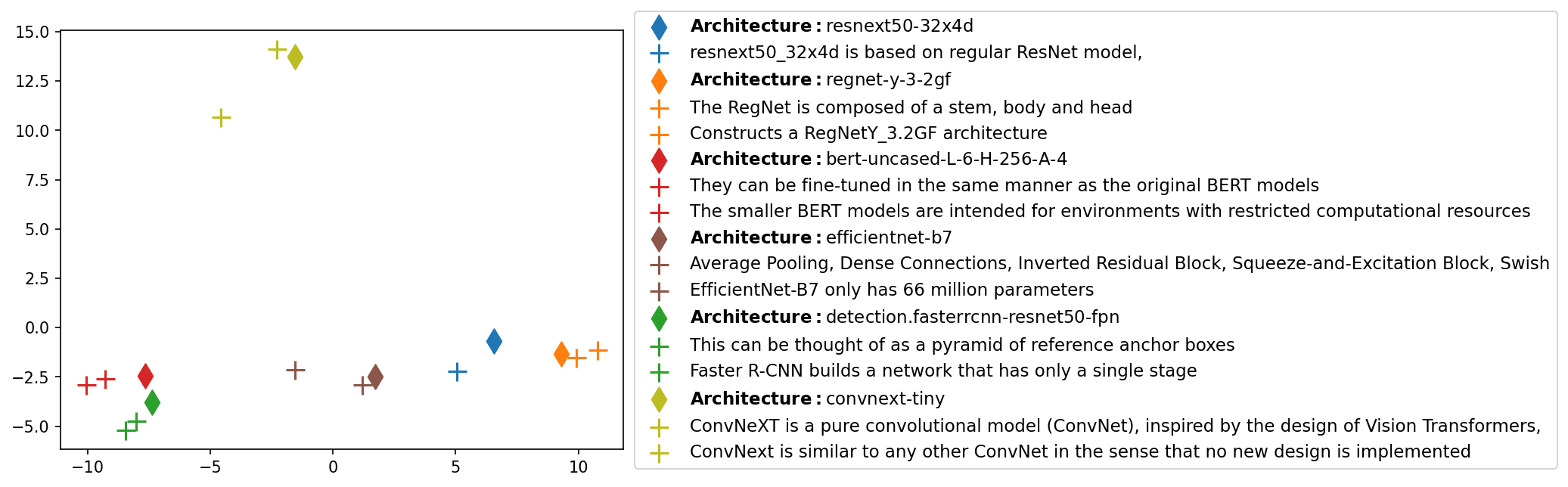}
    \vspace{-15pt}
    \caption{\label{fig:visualization} 
    Visualization of example relevant architecture and text embeddings in a 2D space (projected via PCA).
    }
\end{figure*}

\subsection{Embeddings Visualization}

As discussed before, ArchBERT learns to minimize the cosine distance between relevant text and architecture embeddings, while maximizing the distance for the irrelevant ones. To convey this concept, we visualize the joint embeddings of example relevant texts and architectures (i.e., $J_t$ and $J_g$ in Equation \ref{eq:cross-encoder}) form TVHF dataset in Figure \ref{fig:visualization}. The points in the figure are obtained by projecting the embeddings to a 2D space via PCA \citep{pca}. As shown in Figure \ref{fig:visualization}, the text embeddings are mapped to the points near by their relevant architectures. This implies that ArchBERT has learned to minimize the distance between the related pairs of texts and architectures (i.e., positive samples) and obtain similar embeddings for them. On the other hand, the points for the irrelevant descriptions and architectures are projected far from each other, which shows the success of ArchBERT in maximizing the distance between unrelated pairs. 

\vspace{-2pt}
\section{Conclusion}
\vspace{-2pt}
In this paper, we proposed ArchBERT, a bi-modal solution for joint learning of neural architectures and natural languages. We also introduced a new pre-training technique called Masked Architecture Modeling (MAM) for a better generalization of ArchBERT. In addition, two new bi-modal benchmark datasets called TVHF and AutoNet were presented on which the proposed model was trained and evaluated for different downstream tasks. Five architecture-language-related tasks and applications were introduced in this work to verify the performance of ArchBERT. This work has opened up new avenues for research in the area of architecture-language joint understanding, particularly the proposed benchmarks. Potential research directions to this work include text-based neural architecture generation and bi-modal learning of languages and other graph-structured modalities such as knowledge graphs and social network graphs.

\bibliography{ref}

\begin{thebibliography}{38}
\expandafter\ifx\csname natexlab\endcsname\relax\def\natexlab#1{#1}\fi

\bibitem[{Akbari et~al.(2021)Akbari, Yuan, Qian, Chuang, Chang, Cui, and
  Gong}]{vatt}
Hassan Akbari, Liangzhe Yuan, Rui Qian, Wei-Hong Chuang, Shih-Fu Chang, Yin
  Cui, and Boqing Gong. 2021.
\newblock {VATT}: Transformers for multimodal self-supervised learning from raw
  video, audio and text.
\newblock \emph{Advances in Neural Information Processing Systems},
  34:24206--24221.

\bibitem[{Akbari et~al.(2022)Akbari, Banitalebi-Dehkordi, and Zhang}]{elang}
Mohammad Akbari, Amin Banitalebi-Dehkordi, and Yong Zhang. 2022.
\newblock E-lang: Energy-based joint inferencing of super and swift language
  models.
\newblock In \emph{Proceedings of the 60th Annual Meeting of the Association
  for Computational Linguistics (Volume 1: Long Papers)}, pages 5229--5244.

\bibitem[{Anderson et~al.(2018)Anderson, He, Buehler, Teney, Johnson, Gould,
  and Zhang}]{anderson2018bottom}
Peter Anderson, Xiaodong He, Chris Buehler, Damien Teney, Mark Johnson, Stephen
  Gould, and Lei Zhang. 2018.
\newblock Bottom-up and top-down attention for image captioning and visual
  question answering.
\newblock In \emph{Proceedings of the IEEE conference on computer vision and
  pattern recognition}, pages 6077--6086.

\bibitem[{Baltru{\v{s}}aitis et~al.(2018)Baltru{\v{s}}aitis, Ahuja, and
  Morency}]{baltruvsaitis2018multimodal}
Tadas Baltru{\v{s}}aitis, Chaitanya Ahuja, and Louis-Philippe Morency. 2018.
\newblock Multimodal machine learning: A survey and taxonomy.
\newblock \emph{IEEE transactions on pattern analysis and machine
  intelligence}, 41(2):423--443.

\bibitem[{Bao et~al.(2022)Bao, Wang, Dong, and Wei}]{vlbeit}
Hangbo Bao, Wenhui Wang, Li~Dong, and Furu Wei. 2022.
\newblock Vl-beit: Generative vision-language pretraining.
\newblock \emph{arXiv preprint arXiv:2206.01127}.

\bibitem[{Brown et~al.(2020)Brown, Mann, Ryder, Subbiah, Kaplan, Dhariwal,
  Neelakantan, Shyam, Sastry, Askell et~al.}]{gpt3}
Tom Brown, Benjamin Mann, Nick Ryder, Melanie Subbiah, Jared~D Kaplan, Prafulla
  Dhariwal, Arvind Neelakantan, Pranav Shyam, Girish Sastry, Amanda Askell,
  et~al. 2020.
\newblock Language models are few-shot learners.
\newblock \emph{Advances in neural information processing systems},
  33:1877--1901.

\bibitem[{Devlin et~al.(2019)Devlin, Chang, Lee, and Toutanova}]{bert}
Jacob Devlin, Ming-Wei Chang, Kenton Lee, and Kristina Toutanova. 2019.
\newblock {BERT}: Pre-training of deep bidirectional transformers for language
  understanding.
\newblock In \emph{Proceedings of NAACL-HLT}, pages 4171--4186.

\bibitem[{Feng et~al.(2020)Feng, Guo, Tang, Duan, Feng, Gong, Shou, Qin, Liu,
  Jiang, and Zhou}]{codebert}
Zhangyin Feng, Daya Guo, Duyu Tang, Nan Duan, Xiaocheng Feng, Ming Gong, Linjun
  Shou, Bing Qin, Ting Liu, Daxin Jiang, and Ming Zhou. 2020.
\newblock Codebert: {A} pre-trained model for programming and natural
  languages.
\newblock In \emph{Findings of the Association for Computational Linguistics:
  {EMNLP} 2020, Online Event, 16-20 November 2020}, volume {EMNLP} 2020 of
  \emph{Findings of {ACL}}, pages 1536--1547. Association for Computational
  Linguistics.

\bibitem[{Fu et~al.(2021)Fu, Li, Gan, Lin, Wang, Wang, and Liu}]{violet}
Tsu-Jui Fu, Linjie Li, Zhe Gan, Kevin Lin, William~Yang Wang, Lijuan Wang, and
  Zicheng Liu. 2021.
\newblock Violet: End-to-end video-language transformers with masked
  visual-token modeling.
\newblock \emph{arXiv preprint arXiv:2111.12681}.

\bibitem[{Google(2022)}]{data_quality}
Google. 2022.
\newblock The size and quality of a data set.
\newblock
  \url{https://developers.google.com/machine-learning/data-prep/construct/collect/data-size-quality},
  Last accessed on 2022-12-14.

\bibitem[{Guo et~al.(2021)Guo, Ren, Lu, Feng, Tang, Liu, Zhou, Duan,
  Svyatkovskiy, Fu, Tufano, Deng, Clement, Drain, Sundaresan, Yin, Jiang, and
  Zhou}]{graphcodebert}
Daya Guo, Shuo Ren, Shuai Lu, Zhangyin Feng, Duyu Tang, Shujie Liu, Long Zhou,
  Nan Duan, Alexey Svyatkovskiy, Shengyu Fu, Michele Tufano, Shao~Kun Deng,
  Colin~B. Clement, Dawn Drain, Neel Sundaresan, Jian Yin, Daxin Jiang, and
  Ming Zhou. 2021.
\newblock Graphcodebert: Pre-training code representations with data flow.
\newblock In \emph{9th International Conference on Learning Representations,
  {ICLR} 2021, Virtual Event, Austria, May 3-7, 2021}. OpenReview.net.

\bibitem[{Henderson et~al.(2017)Henderson, Al-Rfou, Strope, Sung, Luk{\'a}cs,
  Guo, Kumar, Miklos, and Kurzweil}]{henderson2017efficient}
Matthew Henderson, Rami Al-Rfou, Brian Strope, Yun-Hsuan Sung, L{\'a}szl{\'o}
  Luk{\'a}cs, Ruiqi Guo, Sanjiv Kumar, Balint Miklos, and Ray Kurzweil. 2017.
\newblock Efficient natural language response suggestion for smart reply.
\newblock \emph{arXiv preprint arXiv:1705.00652}.

\bibitem[{Hong et~al.(2020)Hong, Gao, Yokoya, Yao, Chanussot, Du, and
  Zhang}]{remote-sense1}
Danfeng Hong, Lianru Gao, Naoto Yokoya, Jing Yao, Jocelyn Chanussot, Qian Du,
  and Bing Zhang. 2020.
\newblock More diverse means better: Multimodal deep learning meets
  remote-sensing imagery classification.
\newblock \emph{IEEE Transactions on Geoscience and Remote Sensing},
  59(5):4340--4354.

\bibitem[{Jolliffe(2005)}]{pca}
Ian Jolliffe. 2005.
\newblock Principal component analysis.
\newblock \emph{Encyclopedia of statistics in behavioral science}.

\bibitem[{Knyazev et~al.(2021)Knyazev, Drozdzal, Taylor, and
  Romero~Soriano}]{ppuda}
Boris Knyazev, Michal Drozdzal, Graham~W Taylor, and Adriana Romero~Soriano.
  2021.
\newblock Parameter prediction for unseen deep architectures.
\newblock \emph{Advances in Neural Information Processing Systems},
  34:29433--29448.

\bibitem[{Krizhevsky et~al.(2017)Krizhevsky, Sutskever, and Hinton}]{alexnet}
Alex Krizhevsky, Ilya Sutskever, and Geoffrey~E Hinton. 2017.
\newblock Imagenet classification with deep convolutional neural networks.
\newblock \emph{Communications of the ACM}, 60(6):84--90.

\bibitem[{Lewis et~al.(2020)Lewis, Liu, Goyal, Ghazvininejad, Mohamed, Levy,
  Stoyanov, and Zettlemoyer}]{bart}
Mike Lewis, Yinhan Liu, Naman Goyal, Marjan Ghazvininejad, Abdelrahman Mohamed,
  Omer Levy, Veselin Stoyanov, and Luke Zettlemoyer. 2020.
\newblock Bart: Denoising sequence-to-sequence pre-training for natural
  language generation, translation, and comprehension.
\newblock In \emph{Proceedings of the 58th Annual Meeting of the Association
  for Computational Linguistics}, pages 7871--7880.

\bibitem[{Li et~al.(2019)Li, Yatskar, Yin, Hsieh, and Chang}]{visualbert}
Liunian~Harold Li, Mark Yatskar, Da~Yin, Cho-Jui Hsieh, and Kai-Wei Chang.
  2019.
\newblock {VisualBERT}: A simple and performant baseline for vision and
  language.
\newblock \emph{arXiv preprint arXiv:1908.03557}.

\bibitem[{Lin(2004)}]{lin2004rouge}
Chin-Yew Lin. 2004.
\newblock Rouge: A package for automatic evaluation of summaries.
\newblock In \emph{Text summarization branches out}, pages 74--81.

\bibitem[{Lu et~al.(2019)Lu, Batra, Parikh, and Lee}]{Vilbert}
Jiasen Lu, Dhruv Batra, Devi Parikh, and Stefan Lee. 2019.
\newblock {ViLBERT}: Pretraining task-agnostic visiolinguistic representations
  for vision-and-language tasks.
\newblock \emph{Advances in neural information processing systems}, 32.

\bibitem[{Maimaitijiang et~al.(2020)Maimaitijiang, Sagan, Sidike, Hartling,
  Esposito, and Fritschi}]{remote-sense2}
Maitiniyazi Maimaitijiang, Vasit Sagan, Paheding Sidike, Sean Hartling, Flavio
  Esposito, and Felix~B Fritschi. 2020.
\newblock Soybean yield prediction from uav using multimodal data fusion and
  deep learning.
\newblock \emph{Remote sensing of environment}, 237:111599.

\bibitem[{Marcel and Rodriguez(2010)}]{torchvision}
S\'{e}bastien Marcel and Yann Rodriguez. 2010.
\newblock Torchvision the machine-vision package of torch.
\newblock In \emph{Proceedings of the 18th ACM International Conference on
  Multimedia}, page 1485–1488. Association for Computing Machinery.

\bibitem[{Michelsanti et~al.(2021)Michelsanti, Tan, Zhang, Xu, Yu, Yu, and
  Jensen}]{audiovisual1}
Daniel Michelsanti, Zheng-Hua Tan, Shi-Xiong Zhang, Yong Xu, Meng Yu, Dong Yu,
  and Jesper Jensen. 2021.
\newblock An overview of deep-learning-based audio-visual speech enhancement
  and separation.
\newblock \emph{IEEE/ACM Transactions on Audio, Speech, and Language
  Processing}, 29:1368--1396.

\bibitem[{OpenAI(2022)}]{chatgpt}
OpenAI. 2022.
\newblock Introducing chatgpt.
\newblock \url{https://openai.com/blog/chatgpt}.

\bibitem[{OpenAI(2023)}]{openai2023gpt4}
OpenAI. 2023.
\newblock \href {http://arxiv.org/abs/2303.08774} {Gpt-4 technical report}.

\bibitem[{Raffel et~al.(2020)Raffel, Shazeer, Roberts, Lee, Narang, Matena,
  Zhou, Li, Liu et~al.}]{t5}
Colin Raffel, Noam Shazeer, Adam Roberts, Katherine Lee, Sharan Narang, Michael
  Matena, Yanqi Zhou, Wei Li, Peter~J Liu, et~al. 2020.
\newblock Exploring the limits of transfer learning with a unified text-to-text
  transformer.
\newblock \emph{J. Mach. Learn. Res.}, 21(140):1--67.

\bibitem[{Reimers and Gurevych(2019)}]{sbert}
Nils Reimers and Iryna Gurevych. 2019.
\newblock Sentence-bert: Sentence embeddings using siamese bert-networks.
\newblock In \emph{Proceedings of the 2019 Conference on Empirical Methods in
  Natural Language Processing and the 9th International Joint Conference on
  Natural Language Processing (EMNLP-IJCNLP)}, pages 3982--3992.

\bibitem[{Ren et~al.(2015)Ren, He, Girshick, and Sun}]{fasterrcnn}
Shaoqing Ren, Kaiming He, Ross Girshick, and Jian Sun. 2015.
\newblock Faster r-cnn: Towards real-time object detection with region proposal
  networks.
\newblock \emph{Advances in neural information processing systems}, 28.

\bibitem[{Santisteban and Tejada-C{\'a}rcamo(2015)}]{jaccard2}
Julio Santisteban and Javier Tejada-C{\'a}rcamo. 2015.
\newblock Unilateral jaccard similarity coefficient.
\newblock In \emph{GSB@ SIGIR}, pages 23--27.

\bibitem[{Schoneveld et~al.(2021)Schoneveld, Othmani, and
  Abdelkawy}]{audiovisual3}
Liam Schoneveld, Alice Othmani, and Hazem Abdelkawy. 2021.
\newblock Leveraging recent advances in deep learning for audio-visual emotion
  recognition.
\newblock \emph{Pattern Recognition Letters}, 146:1--7.

\bibitem[{Sun et~al.(2019)Sun, Myers, Vondrick, Murphy, and Schmid}]{videobert}
Chen Sun, Austin Myers, Carl Vondrick, Kevin Murphy, and Cordelia Schmid. 2019.
\newblock {VideoBERT}: A joint model for video and language representation
  learning.
\newblock In \emph{Proceedings of the IEEE/CVF International Conference on
  Computer Vision}, pages 7464--7473.

\bibitem[{Tan and Bansal(2019)}]{LXMERT}
Hao Tan and Mohit Bansal. 2019.
\newblock {LXMERT:} learning cross-modality encoder representations from
  transformers.
\newblock In \emph{Proceedings of the 2019 Conference on Empirical Methods in
  Natural Language Processing and the 9th International Joint Conference on
  Natural Language Processing, {EMNLP-IJCNLP} 2019, Hong Kong, China, November
  3-7, 2019}, pages 5099--5110. Association for Computational Linguistics.

\bibitem[{Tsai et~al.(2019)Tsai, Bai, Liang, Kolter, Morency, and
  Salakhutdinov}]{unaligned-multimodal}
Yao-Hung~Hubert Tsai, Shaojie Bai, Paul~Pu Liang, J~Zico Kolter, Louis-Philippe
  Morency, and Ruslan Salakhutdinov. 2019.
\newblock Multimodal transformer for unaligned multimodal language sequences.
\newblock In \emph{Proceedings of the conference. Association for Computational
  Linguistics. Meeting}, volume 2019, page 6558. NIH Public Access.

\bibitem[{Vale-Silva and Rohr(2021)}]{medical-multimodal2}
Lu{\'\i}s~A Vale-Silva and Karl Rohr. 2021.
\newblock Long-term cancer survival prediction using multimodal deep learning.
\newblock \emph{Scientific Reports}, 11(1):1--12.

\bibitem[{Veli{\v{c}}kovi{\'c} et~al.(2018)Veli{\v{c}}kovi{\'c}, Cucurull,
  Casanova, Romero, Li{\`o}, and Bengio}]{gat}
Petar Veli{\v{c}}kovi{\'c}, Guillem Cucurull, Arantxa Casanova, Adriana Romero,
  Pietro Li{\`o}, and Yoshua Bengio. 2018.
\newblock Graph attention networks.
\newblock In \emph{International Conference on Learning Representations}.

\bibitem[{Venugopalan et~al.(2021)Venugopalan, Tong, Hassanzadeh, and
  Wang}]{medical-multimodal1}
Janani Venugopalan, Li~Tong, Hamid~Reza Hassanzadeh, and May~D Wang. 2021.
\newblock Multimodal deep learning models for early detection of alzheimer’s
  disease stage.
\newblock \emph{Scientific reports}, 11(1):1--13.

\bibitem[{Wolf et~al.(2019)Wolf, Debut, Sanh, Chaumond, Delangue, Moi, Cistac,
  Rault, Louf, Funtowicz et~al.}]{huggingface}
Thomas Wolf, Lysandre Debut, Victor Sanh, Julien Chaumond, Clement Delangue,
  Anthony Moi, Pierric Cistac, Tim Rault, R{\'e}mi Louf, Morgan Funtowicz,
  et~al. 2019.
\newblock Huggingface's transformers: State-of-the-art natural language
  processing.
\newblock \emph{arXiv preprint arXiv:1910.03771}.

\bibitem[{Xiao et~al.(2020)Xiao, Codevilla, Gurram, Urfalioglu, and
  L{\'o}pez}]{AV1}
Yi~Xiao, Felipe Codevilla, Akhil Gurram, Onay Urfalioglu, and Antonio~M
  L{\'o}pez. 2020.
\newblock Multimodal end-to-end autonomous driving.
\newblock \emph{IEEE Transactions on Intelligent Transportation Systems}.

\end{thebibliography}
\bibliographystyle{acl_natbib}

\clearpage
\appendix
\section{Appendix}
\label{sec:appendix}
\subsection{Code, Dataset, and Demo}
In order for the results to be reproducible, we share our test code (plus the pre-trained model files) with detailed instructions in the supplementary materials. The code also includes the scripts for generating both TVHF and AutoNet datasets.

We also uploaded 6 video files demonstrating the performance of ArchBERT on the following downstream tasks: architecture search (AS), architectural reasoning (AR), architecture clone detection (ACD), bi-modal architecture clone detection (BACD), architectural question answering (AQA), and architecture captioning (AC).

All the code and demo files are also available \href{https://developer.huaweicloud.com/develop/aigallery/notebook/detail?id=e6a924c7-735a-4e02-a25b-4416b77b6315}{here}.

BACD task is similar to ACD, except that a supporting text, which is considered as an extra criteria to refine the results, is also provided along with the two given architectures. The average similarity of the architectures’ embeddings with the help of the text embeddings is evaluated to check if the architectures are similar or not.

\begin{figure*}[!b]
    \centering
    \includegraphics[width=0.99\linewidth]{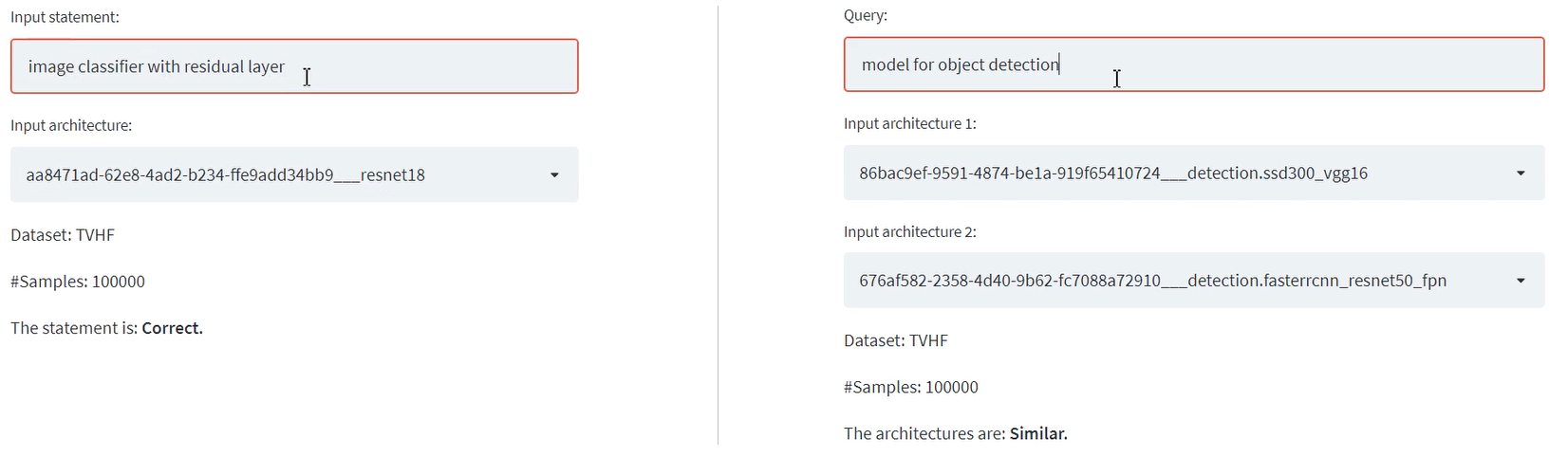}
    \caption{Screenshots from the demos. Left: Architectural Reasoning (AR); Right: Bi-Modal Architecture Clone Detection (BACD).}
    \label{supp:demo_screenshot}
\end{figure*}

The video recordings were taken from a web application we built to demonstrate the real-world application of our method. Example screenshots of the AR and BACD demos are shown in Figure \ref{supp:demo_screenshot}.

\subsection{ArchBERT's Performance on OOD Data}
In order to study the behaviour of ArchBERT on out-of-distribution (OOD) data, we establish another set of experiments on individual TV and HF datasets that have different distributions. In this regard, we pre-train ArchBERT on each of TVHF, TV-only, and HF-only datasets, and evaluate their performance on each other. The corresponding experimental results are summarized in Table \ref{tbl:results-subset}. 

\begin{figure*}
    \includegraphics[width=.31\textwidth]{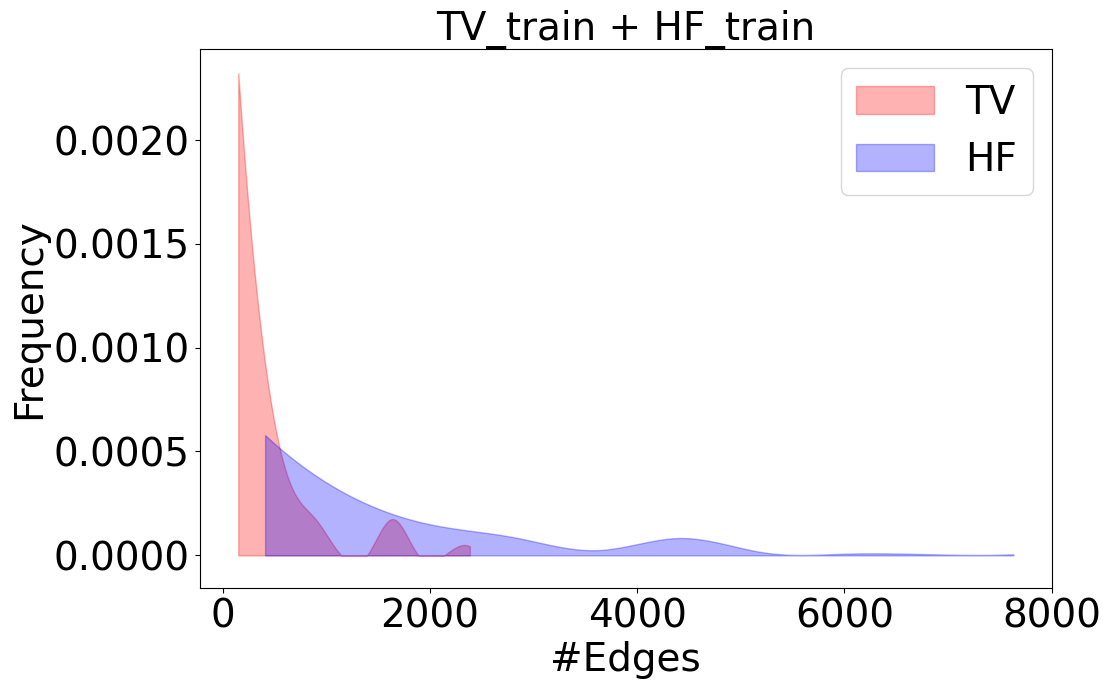}\hfill
    \includegraphics[width=.31\textwidth]{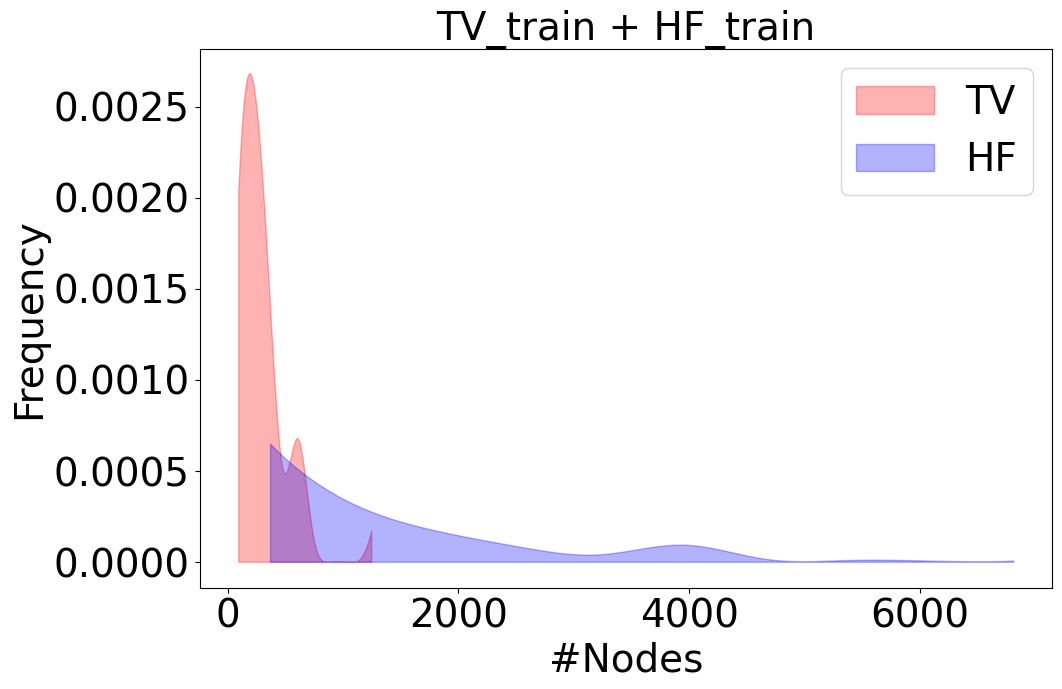}\hfill
    \includegraphics[width=.31\textwidth]{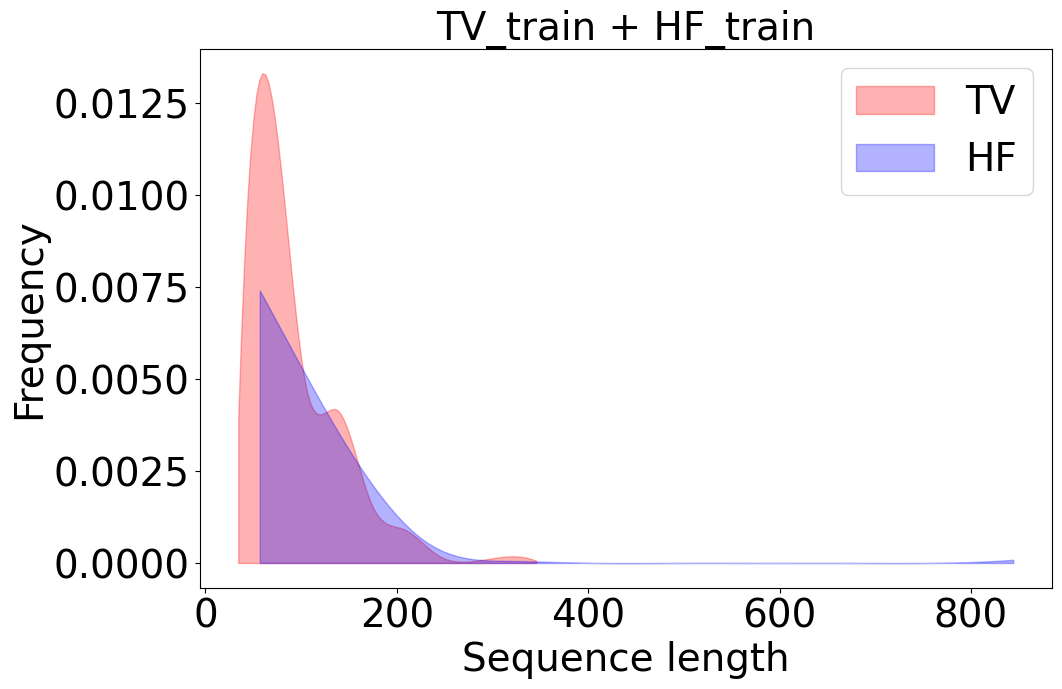}\hfill
    \\[\smallskipamount]
    \includegraphics[width=.31\textwidth]{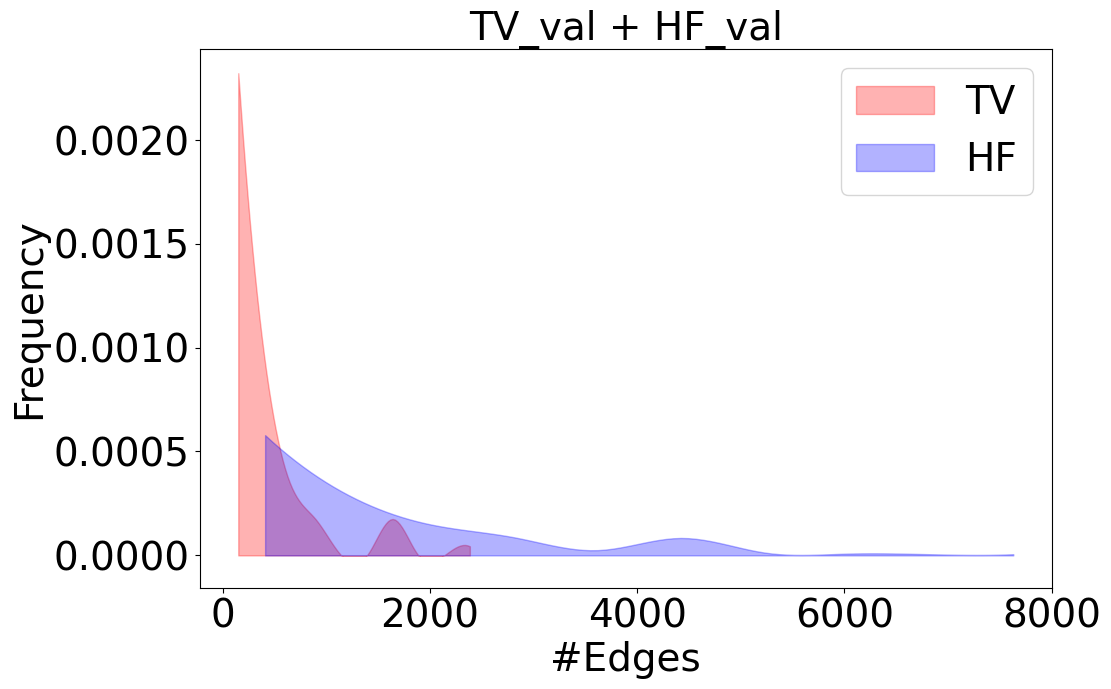}\hfill
    \includegraphics[width=.31\textwidth]{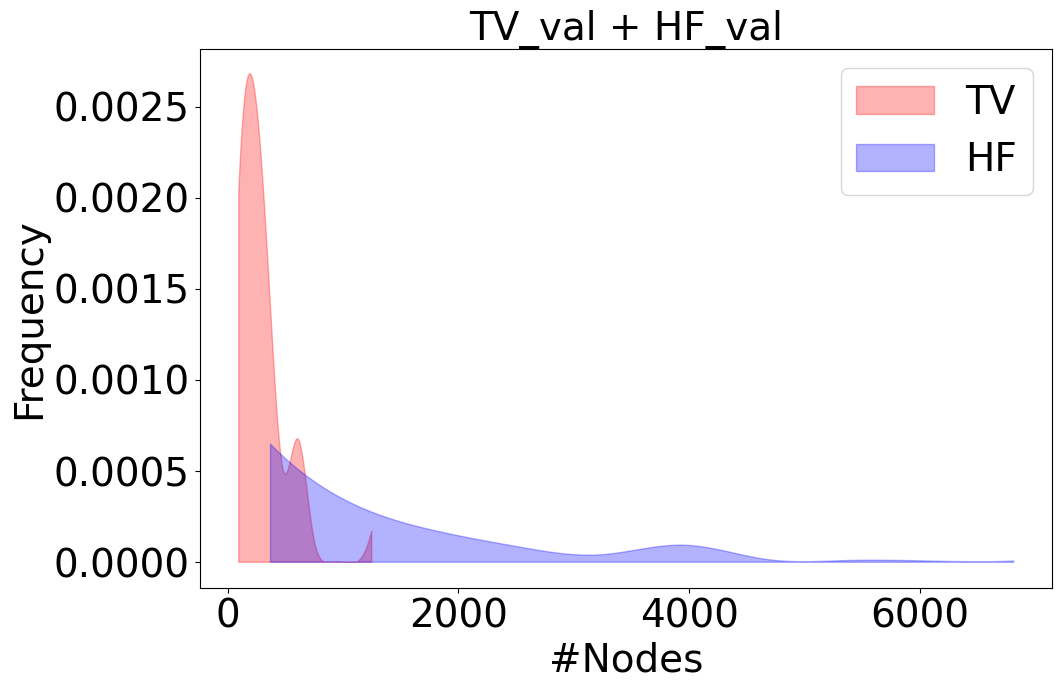}\hfill
    \includegraphics[width=.31\textwidth]{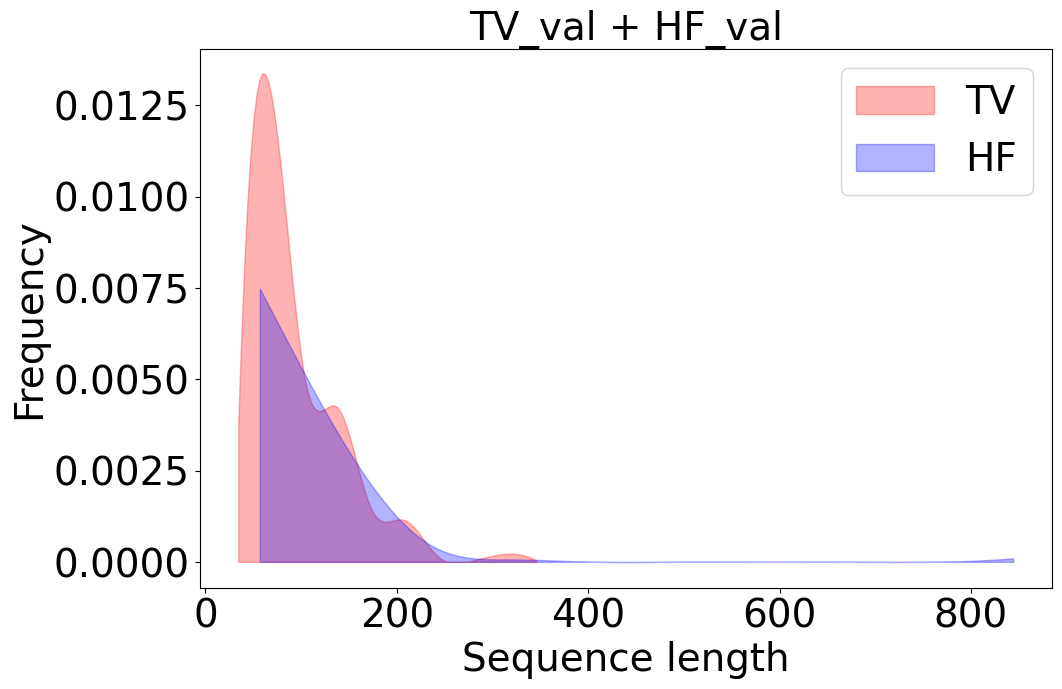}\hfill    
    \caption{Distribution plots of TV and HF train and validation sub-datasets compared with each other.}\label{fig:distplots-ood}
\end{figure*}

As observed in the table, the models trained on TV and HF subsets do not generalize to each other due to the difference in their data distributions, which results in poor performance. The distribution plots for TV and HF subsets are shown in Figure \ref{fig:distplots-ood}. As given in Table \ref{tbl:results-subset}, the highest scores on each of TV and HF subsets are obtained by the model trained with the entire TVHF training dataset. In order to improve the performance of our model on OOD, some techniques such as zero-shot or few-shot learning can be employed, which is a potential research direction for this work. 

\subsection{Embeddings Visualization}
\begin{figure}
  \centering
  \begin{subfigure}[t]{\linewidth}
    \centering\includegraphics[width=\linewidth]{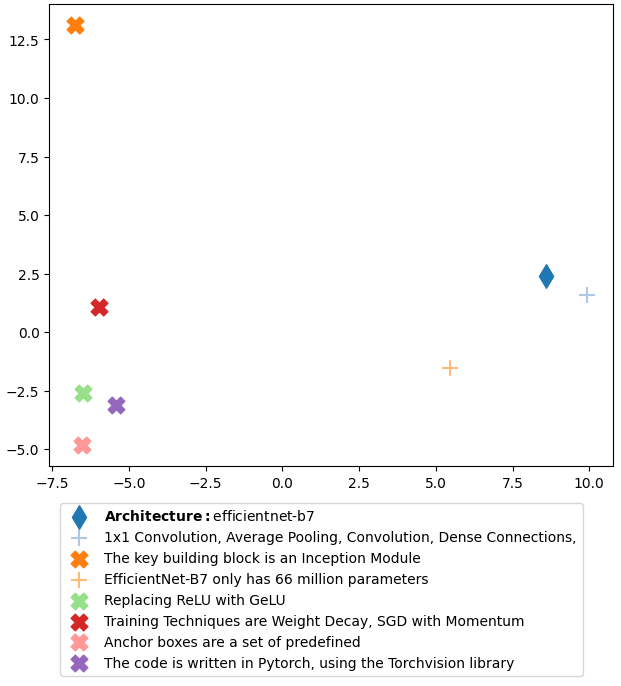}
    \caption{}
  \end{subfigure}
  \\
  \begin{subfigure}[t]{\linewidth}
    \centering\includegraphics[width=\linewidth]{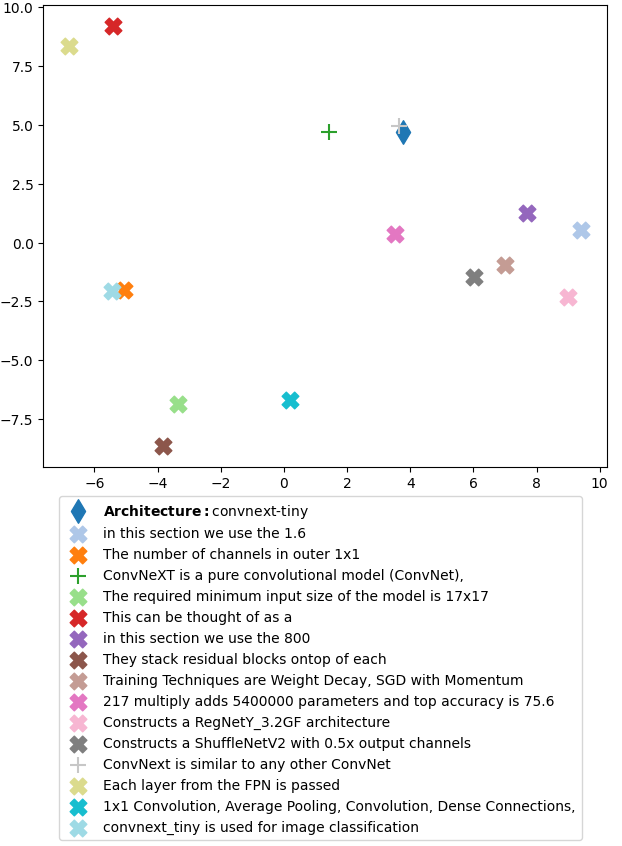}
    \caption{}
  \end{subfigure}
    \caption{Visualization of example pairs of (ir)relevant architecture and text embeddings in a 2D space (projected via PCA).}\label{fig:negative_samples}
\end{figure}

In Figure \ref{fig:visualization}, an embedding visualization of some architecture-text pairs was illustrated. In Figure \ref{fig:negative_samples}, the visualizations for two different architectures from TVHF dataset are individually presented. The points on the figures are obtained by projecting the final ArchBERT's embeddings onto a 2D space via PCA. As shown in the plots, unlike the relevant text embeddings (marked with $+$), the irrelevant ones (marked with $\times$) are projected far from the corresponding architecture embedddings.

\begin{table}[!tb]
\begin{center}
\begin{tabular}[t]{p{1.3cm}p{1.1cm}p{1.1cm}p{1.1cm}p{1.1cm}}
\toprule
\hspace{-5pt}   &  & \multicolumn{3}{c}{\textbf{F1 on Validation set}}  \\\cline{3-5}
\hspace{-5pt}\textbf{Train set} & \textbf{Task} & \textbf{TV}	& \textbf{HF} &	\textbf{TVHF}\\
\\[-0.35cm]
\midrule
\hspace{-5pt}  \multirow{2}{*}{TV} & {AR} &85.05& 3.82&	28.78
\ \\
\hspace{-5pt} & ACD & 	58.88&	22.85&	23.30
 \\
\midrule
\hspace{-5pt}  \multirow{2}{*}{HF} & {AR} & 9.19 &	64.26&	42.43\\
\hspace{-5pt} & ACD & 15.42&	59.98&	54.57
	\\
\midrule
\hspace{-5pt}  \multirow{2}{*}{TVHF} & {AR} &\textbf{85.32}&\textbf{64.39}&	\textbf{71.86}
	\\
\hspace{-5pt} &ACD & \textbf{62.77}&	\textbf{60.0}1&	\textbf{60.10}
 \\
\bottomrule
\end{tabular}
\end{center}
\caption{\label{tbl:results-subset} ArchBERT's performance on OOD data.}
\vspace{10pt}
\end{table}

\subsection{Data Generation}

The procedure of creating TVHF dataset along with negative samples are given in Algorithm \ref{alg:TVHF}. To generate the negative data samples, a pre-trained S-BERT model \citep{sbert} is used to calculate the similarity score between all possible pairs of unique descriptions. If the maximum similarity score between each unique sentence and all other sentences of each unique neural architecture is smaller than a threshold 0.5, that sentence is chosen as an irrelevant description for that specific neural architecture. Note that 93\% of the final TVHF train set contains negative samples. {The above-mentioned procedure of generating many negative candidates per each positive sample was inspired by the multiple negatives sampling idea described by \citet{henderson2017efficient}. Having multiple negatives was proved to be effective when used with dot-product and cosine similarity loss function (Equation \ref{eq:cos_loss} in the main paper).}

For TVHF-ACD dataset, all possible pairs of neural architectures were compared based on their structures. A hard score of 1 or 0 is then assigned to a similar or dissimilar pair of architectures, respectively. For TorchVision architectures with the same architectural base (e.g., ResNet family), a hard score of 1 is assigned to the pair. For HuggingFace models, the configuration files were compared and in case of having similar specifications, a hard score of 1 has been assigned to those architectures. In overall, the TVHF-ACD dataset includes 11\% of similar pairs of architectures.

For AutoNet dataset, all unique layers of each architecture are first extracted. To do so, an algorithm is developed to take an architecture as input and recursively extracts all unique modules and their class path within that architecture. These unique layers are then used along with a list of various pre-defined templates to randomly generate meaningful descriptions with different words and sentence structures. The algorithm is then used with modules that are not included in the architecture to generate irrelevant descriptions that are considered as negative data samples. Each architecture has about 10-11 different descriptions about 30\% of which are the positive ones. The same extracted layers and procedures are also used for automatically generating the question and answer pairs, but with a different set of templates for questions.

\begin{algorithm}
\begin{algorithmic}
  \caption{TVHF dataset generator}
  \label{alg:TVHF}
    \STATE \textbf{Input}: Threshold $\beta$, architectures $G$, pos\_samples $T^p$
    \STATE \textbf{Output}: list of architectures plus their positive and negative descriptions
    \FOR{each unique neural architecture $G_j \in G$}
        \FOR{each unique description $T^p_i \in T^p (G_j)$}
            \IF{max(SBERT($T^p_i$, $T^p_{\sim i}$))  $\leq$ $\beta$ }
            \STATE Add $T_i$ to $T^n (G_j)$ (list of neg\_samples for $j$th architecture)
            \ENDIF
        \ENDFOR
    \ENDFOR
    \RETURN \{$G$, ($T^p$, $T^n$)\}
\end{algorithmic}
\end{algorithm}

\subsection{Distribution Plots for TVHF and AutoNet}
\begin{figure*}
    \includegraphics[width=.31\textwidth]{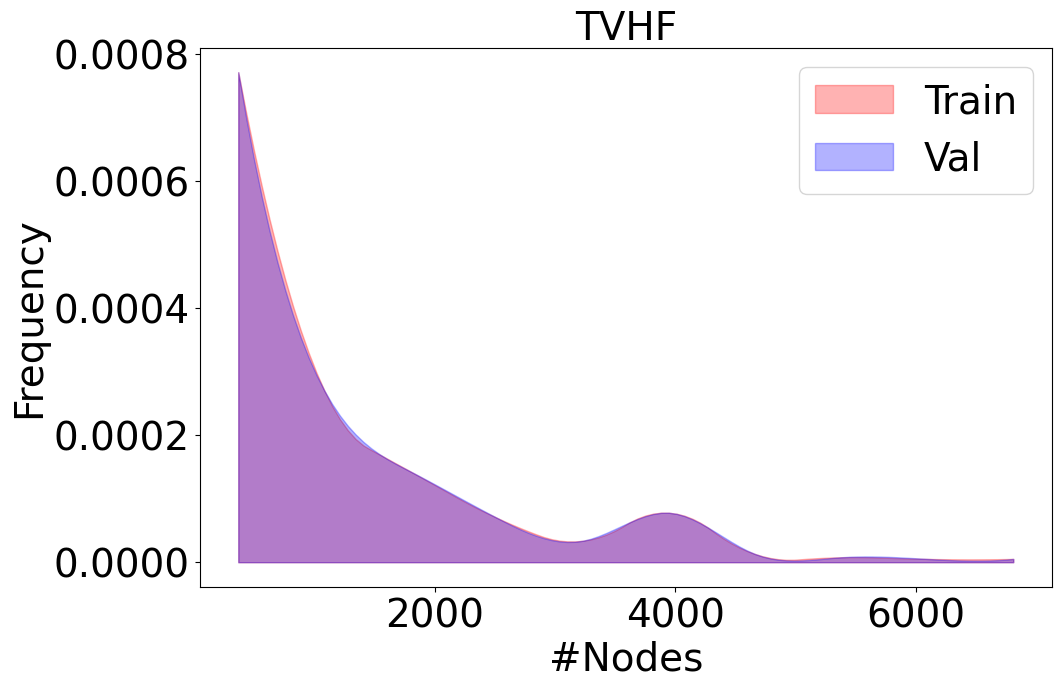}\hfill
    \includegraphics[width=.31\textwidth]{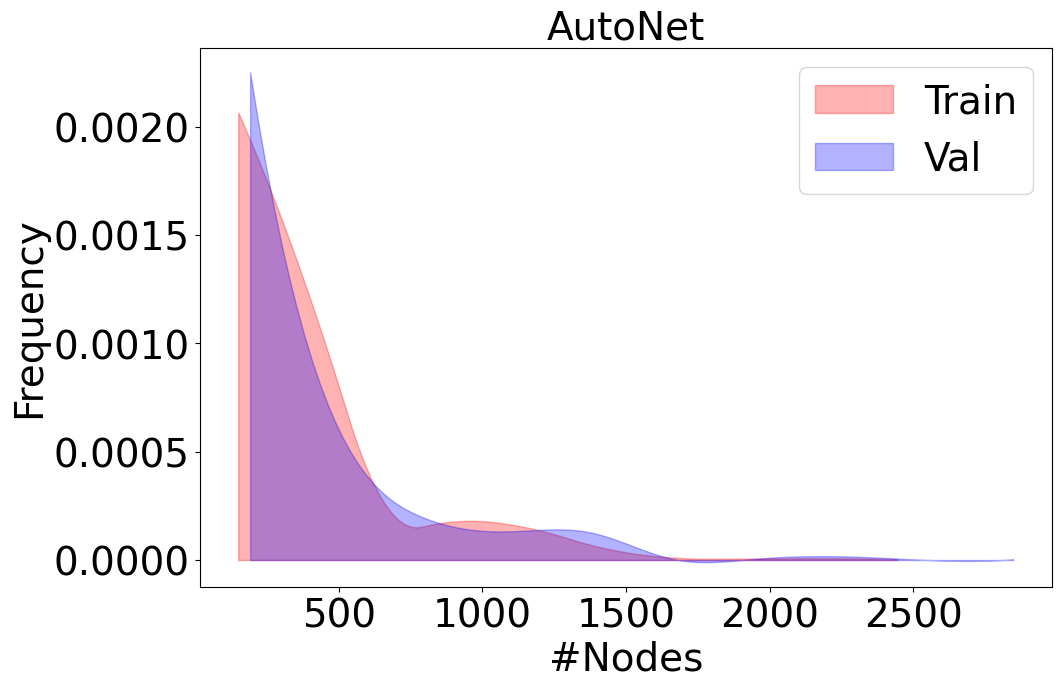}\hfill
    \includegraphics[width=.31\textwidth]{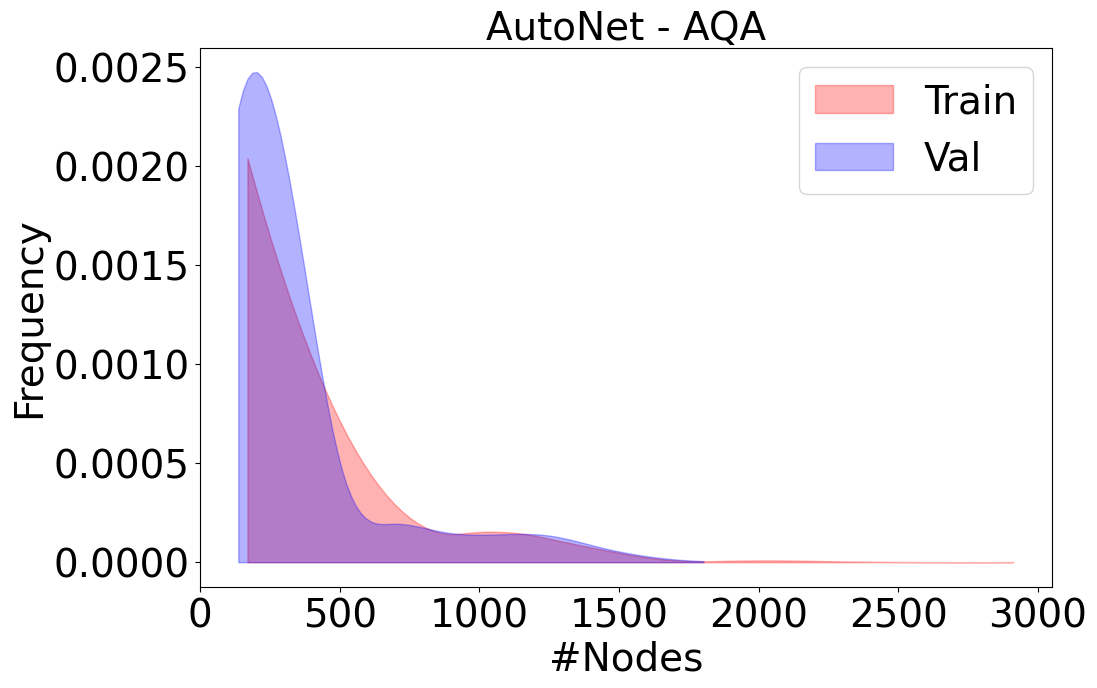}
    \\[\smallskipamount]
    \includegraphics[width=.31\textwidth]{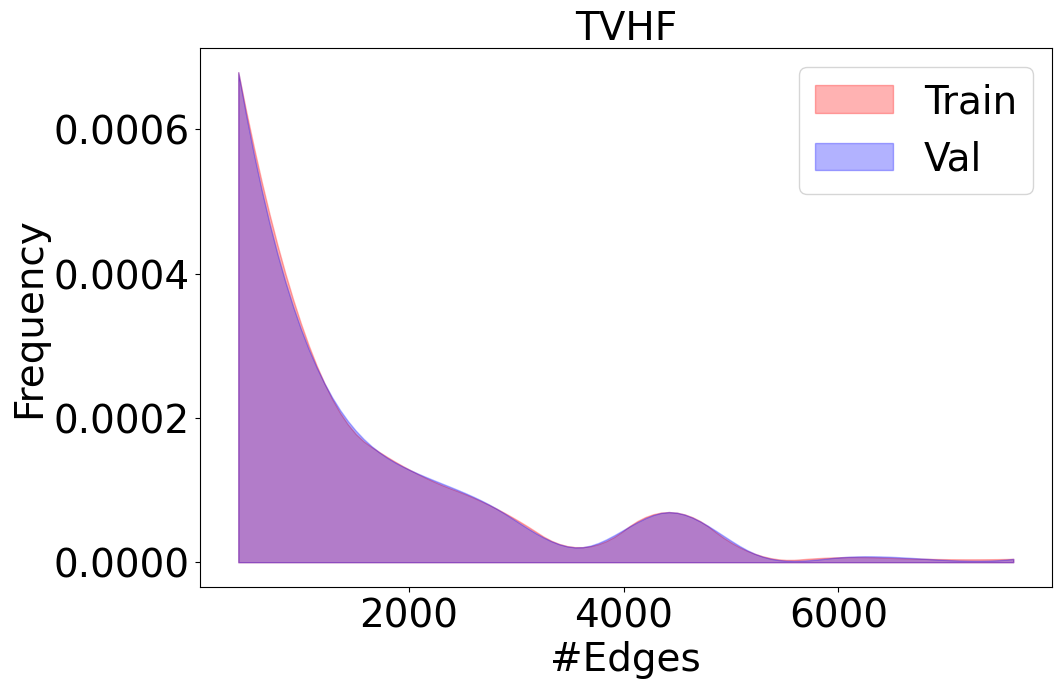}\hfill
    \includegraphics[width=.31\textwidth]{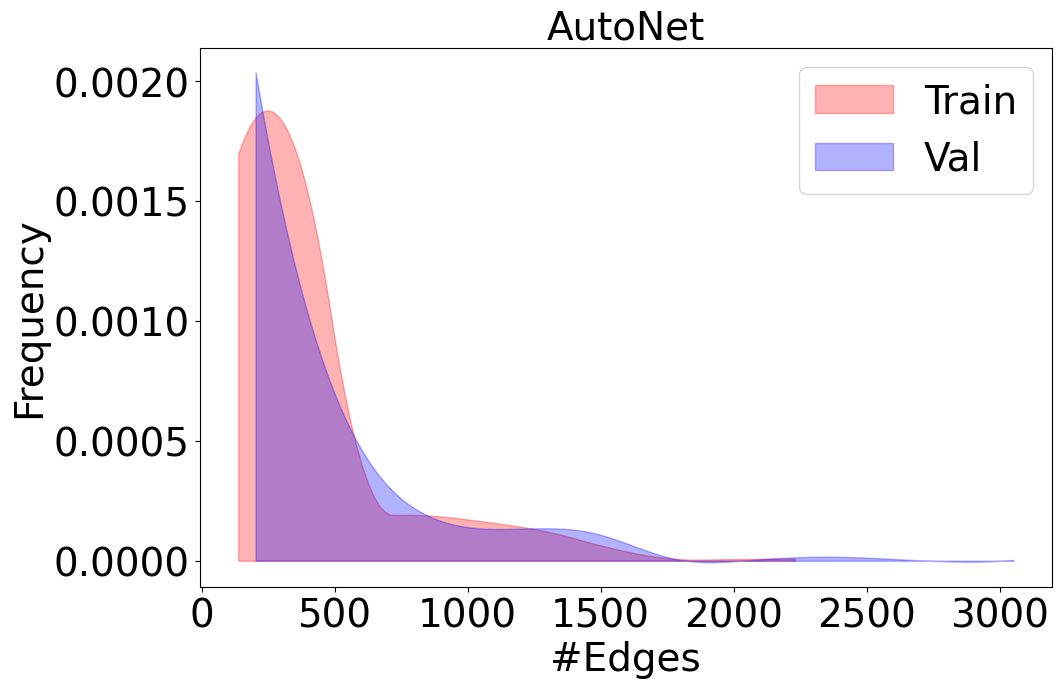}\hfill
    \includegraphics[width=.31\textwidth]{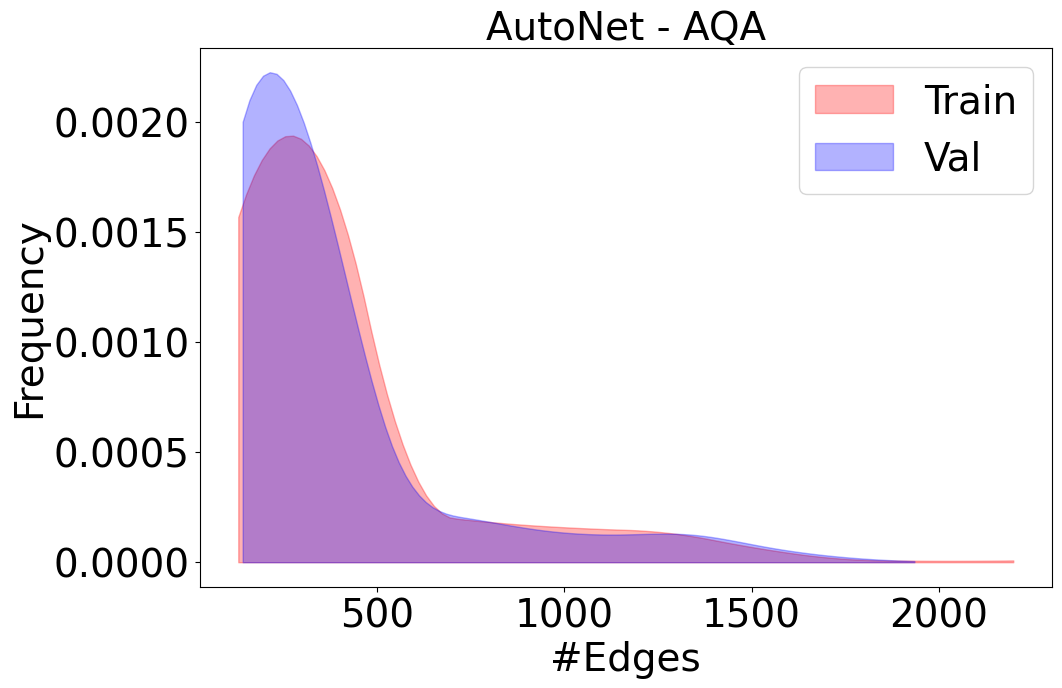}
    \\[\smallskipamount]
    \includegraphics[width=.31\textwidth]{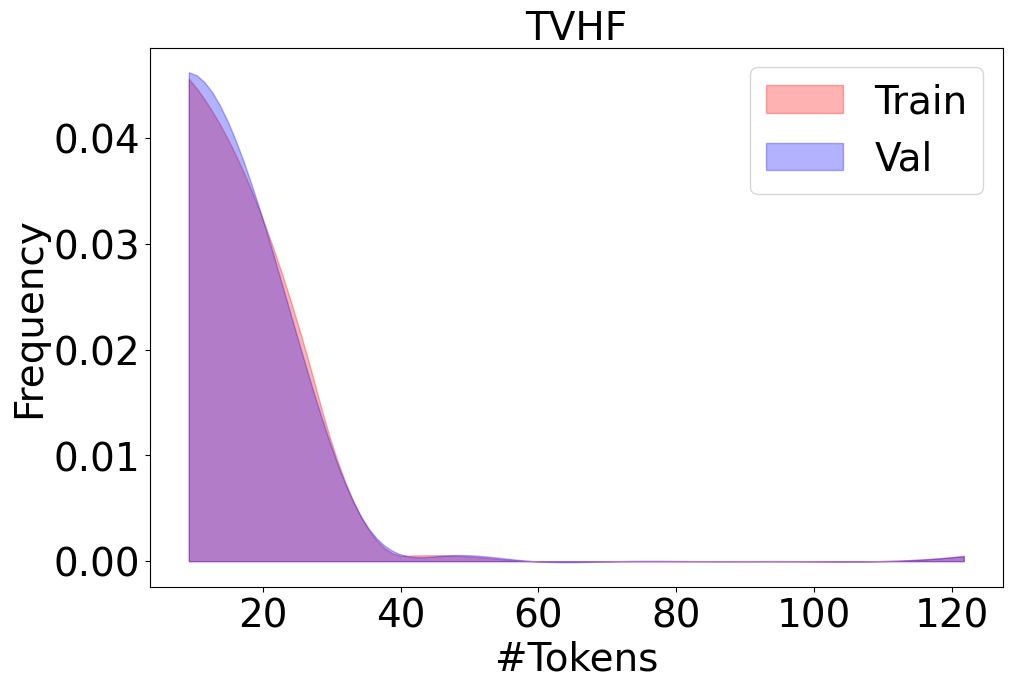}\hfill
    \includegraphics[width=.31\textwidth]{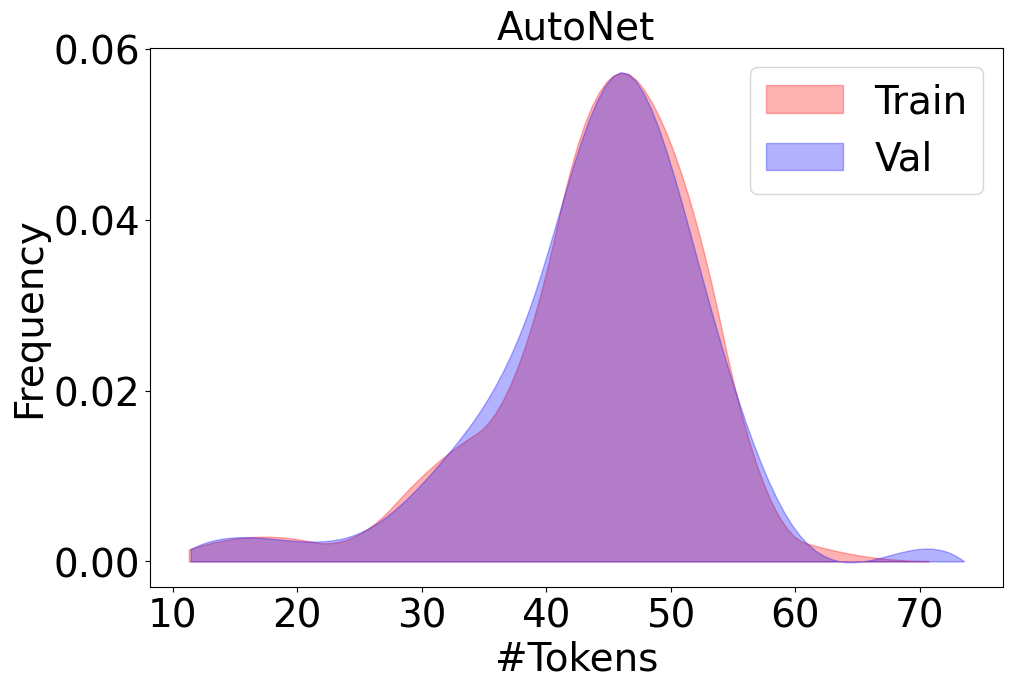}\hfill
    \includegraphics[width=.31\textwidth]{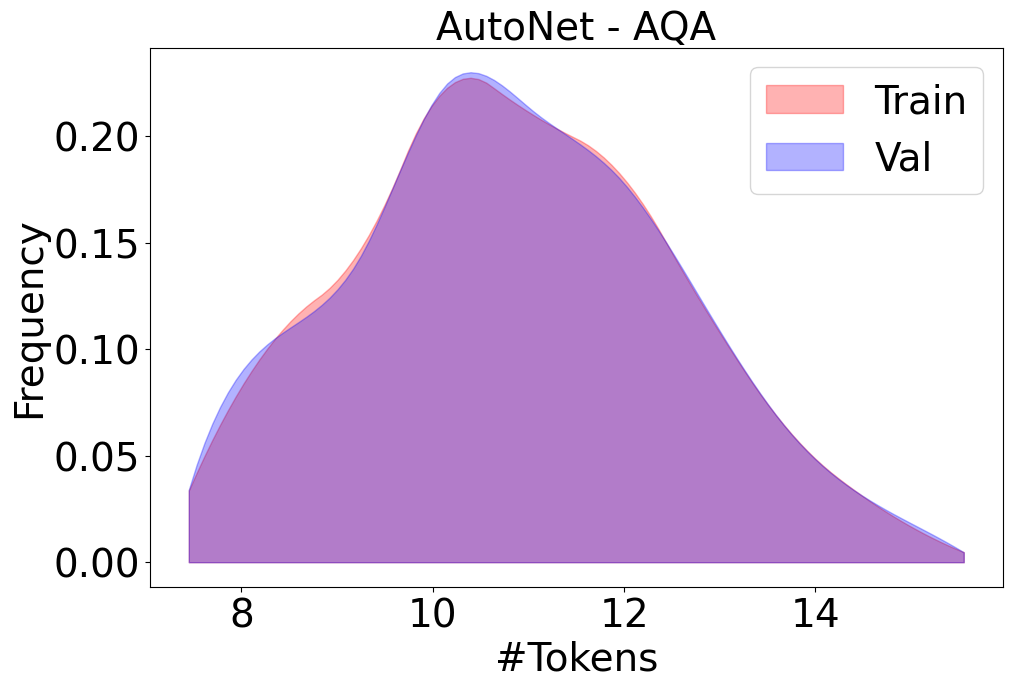}
     \\[\smallskipamount]
     \includegraphics[width=.31\textwidth]{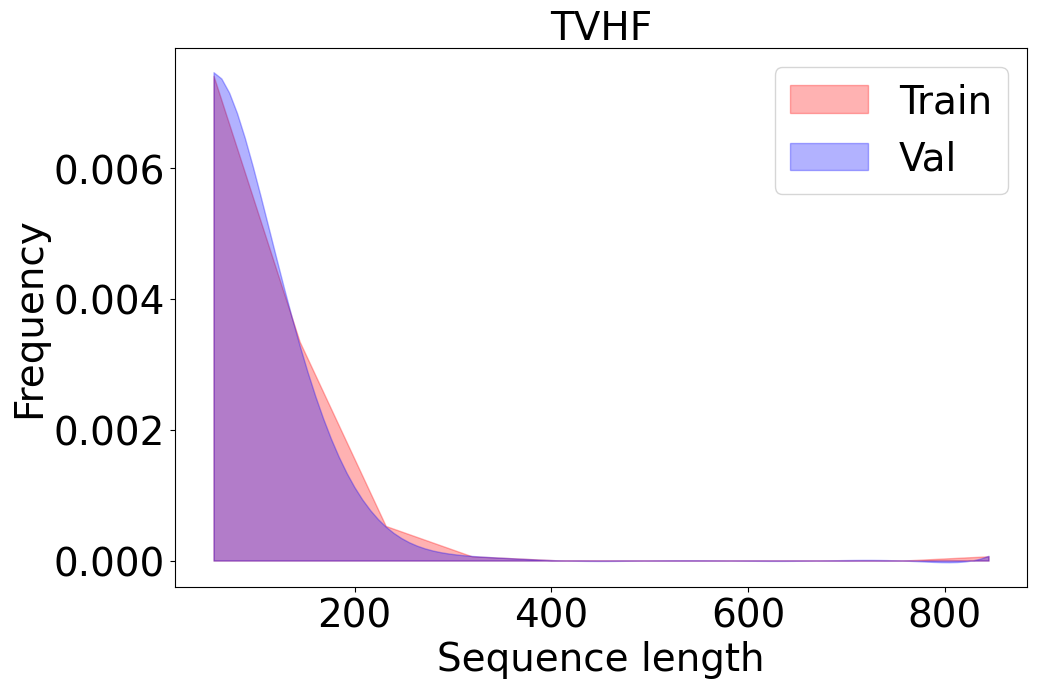}\hfill
     \includegraphics[width=.31\textwidth]{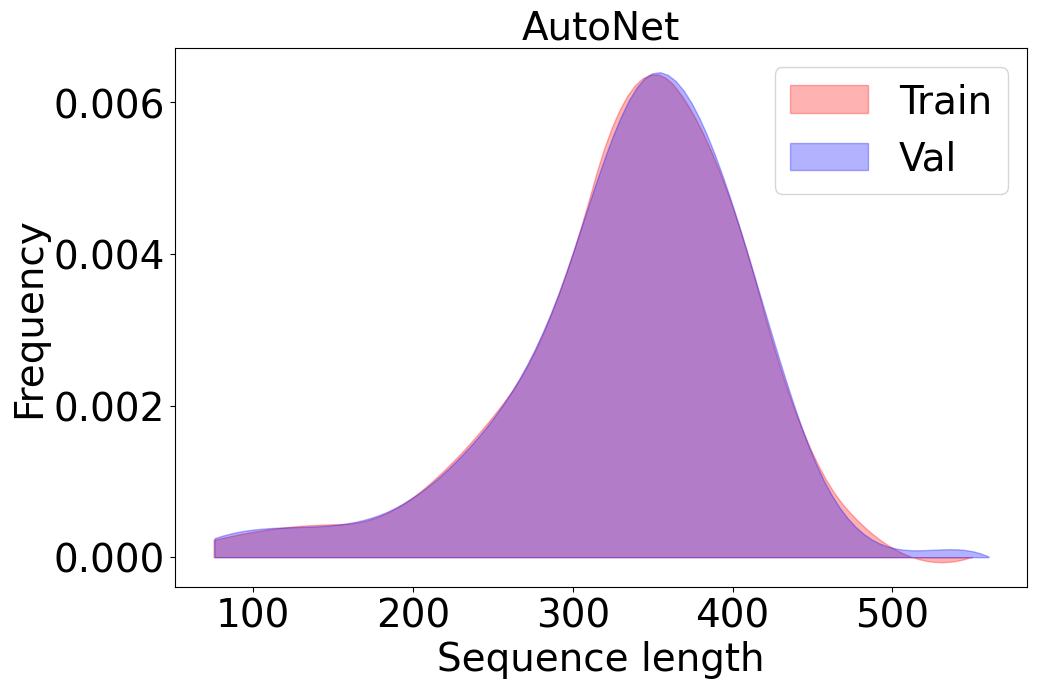}\hfill
     \includegraphics[width=.31\textwidth]{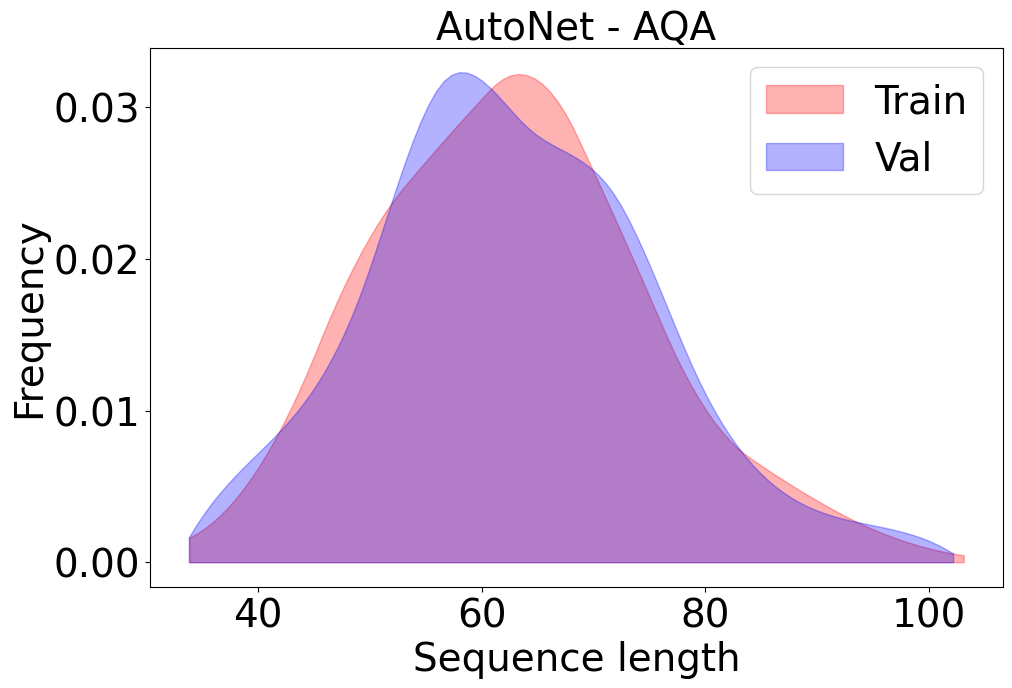}
    \caption{Distribution plots of TVHF, AutoNet, and AutoNet-AQA train/validation datasets.}\label{fig:distplots}
\end{figure*}

Figure \ref{fig:distplots} shows the distribution plots of the TVHF, AutoNet, and AutoNet-AQA datasets. For each dataset, the plots of the training and validation distributions of the number of nodes, the number of edges, the number of textual tokens, and the sequence length of the descriptions are illustrated.

\subsection{Sample Data from TVHF and AutoNet}
In Table \ref{tbl:datasets-TVHF}, example positive architecture-description pairs (for both computer vision and natural language processing problems) from TVHF dataset are given.

Some sample pairs of architectures (with their corresponding "similar" or "dissimilar" ground truth labels) from TVHF-ACD dataset are also presented in Table \ref{tbl:datasets-CD}.

In Table \ref{tbl:datasets-BCD}, we also provide data samples for the BACD task, which includes quartets of two architectures, supporting description, and the similarity label. Note that the numerical analysis of ArchBERT over BACD is not provided because our BACD validation dataset is not finalized to be used for this matter. 

Table \ref{tbl:datasets-AutoNet} also presents a few data samples from AutoNet dataset used for fine-tuning and evaluating ArchBERT on AC task. In Table \ref{tbl:datasets-AQA}, sample data from AutoNet-AQA including the automatically generated questions and ground truth answers  for AQA downstream task are given.

\begin{figure*}
  \centering

  \begin{subfigure}[h!]{.3\textwidth}
    \centering\includegraphics[width=\textwidth]{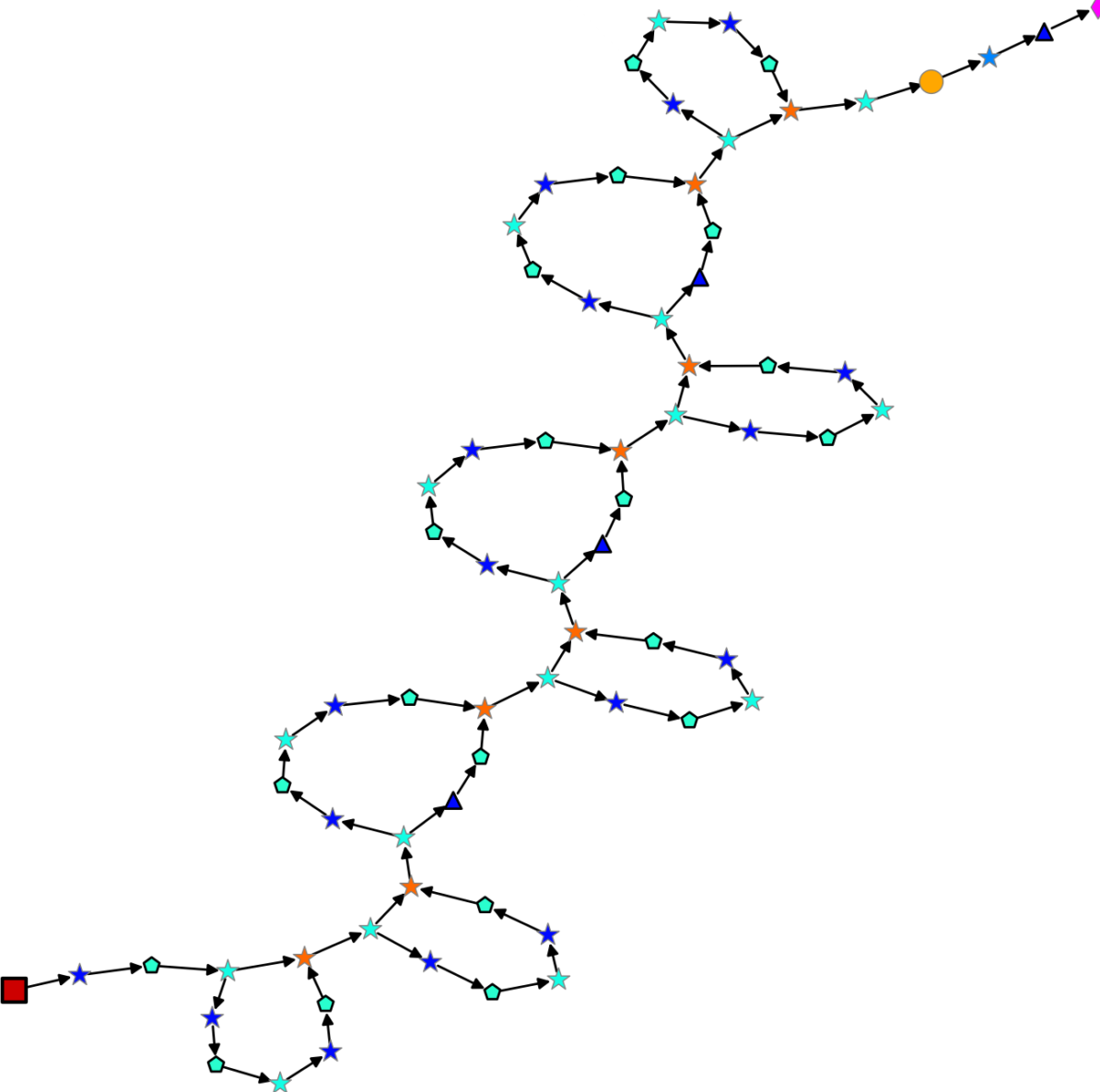}
    \caption{ResNet18}
  \end{subfigure}
  \begin{subfigure}[h!]{.3\textwidth}
    \centering\includegraphics[width=\textwidth]{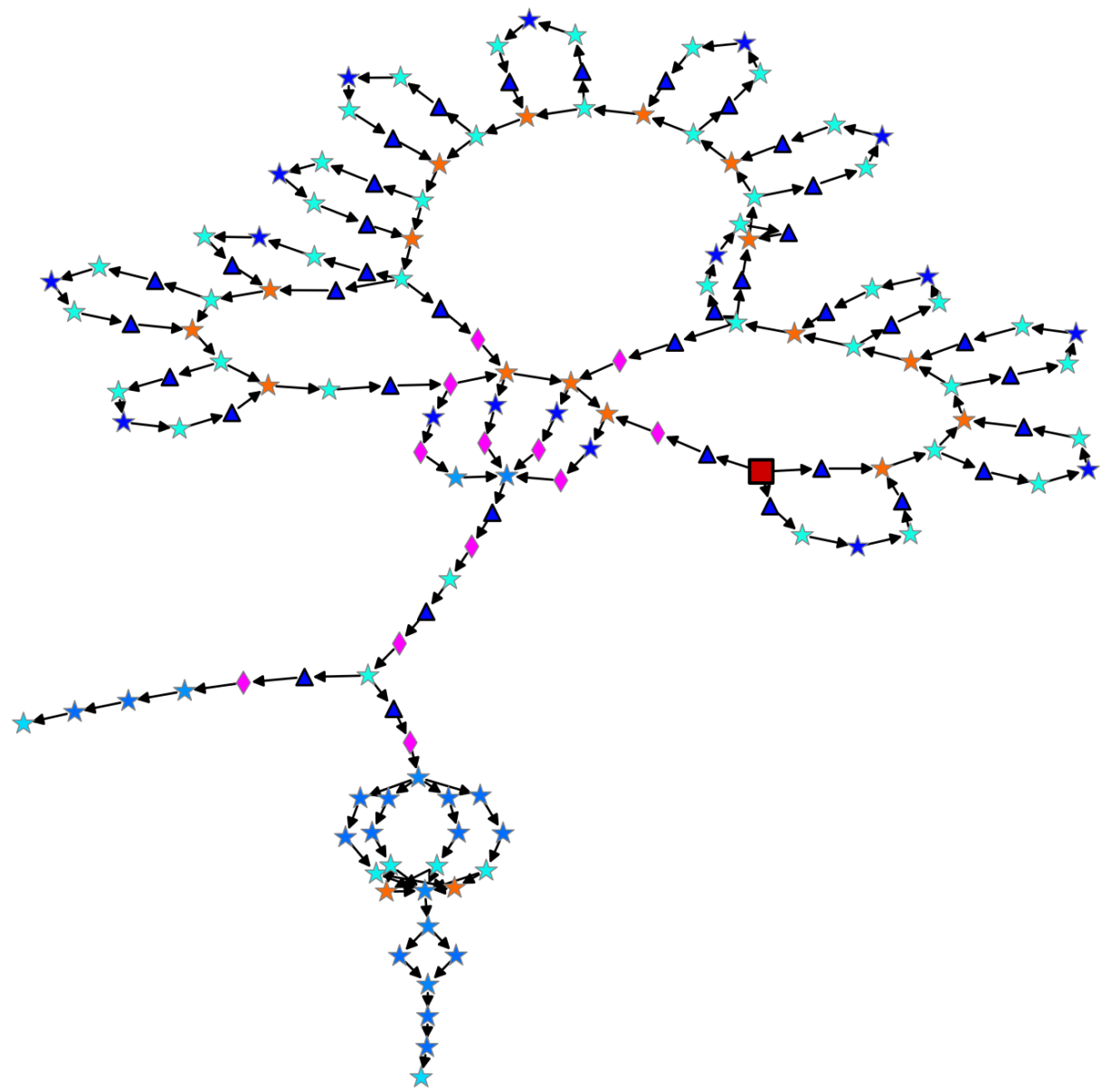}
    \caption{Fasterrcnn-ResNet50-FPN}
  \end{subfigure}
  \begin{subfigure}[h!]{.3\textwidth}
    \centering\includegraphics[width=\textwidth]{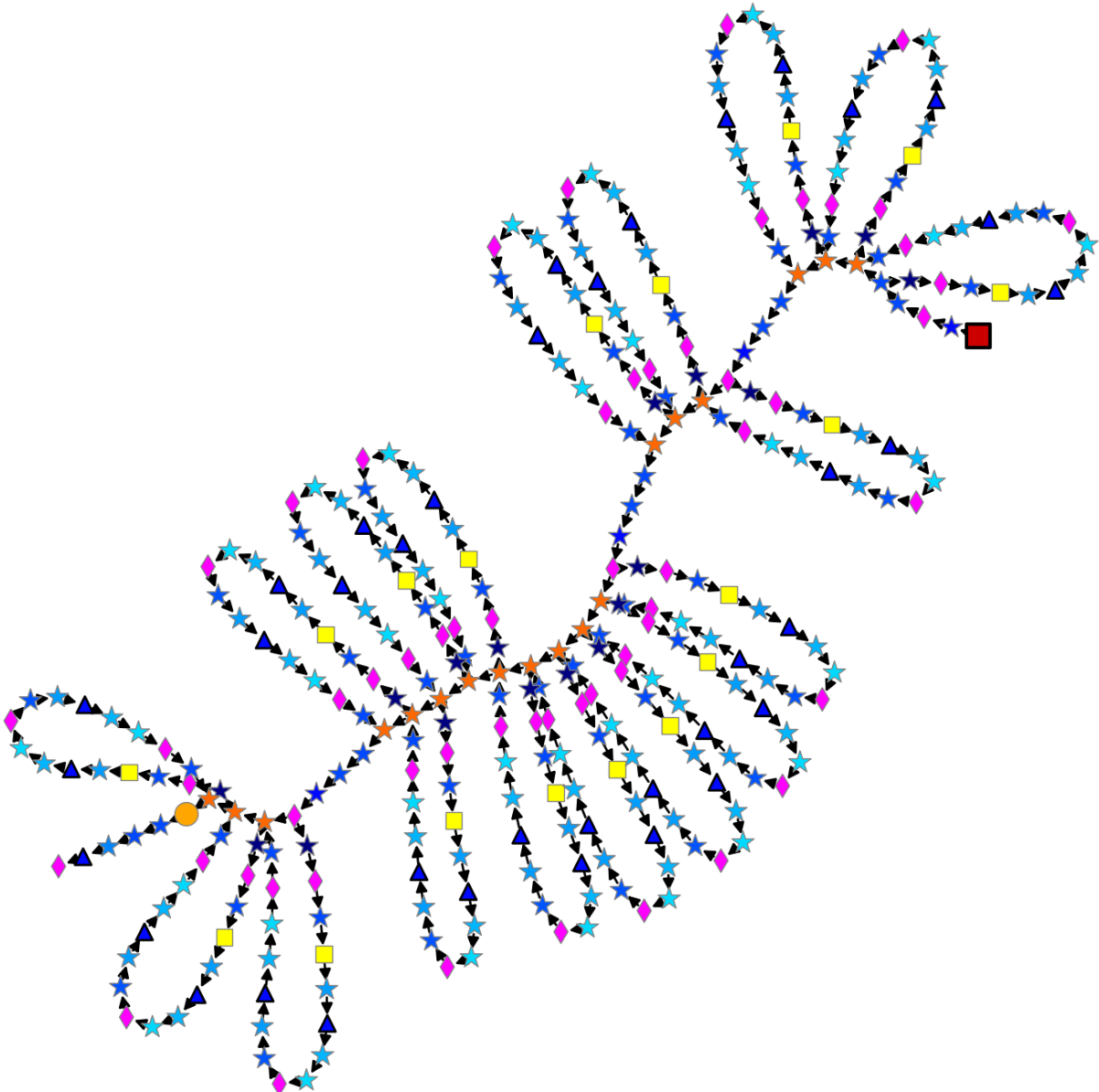}
    \caption{ConvNext-tiny}
  \end{subfigure}
  \begin{subfigure}[h!]{.3\textwidth}
    \centering\includegraphics[width=\textwidth]{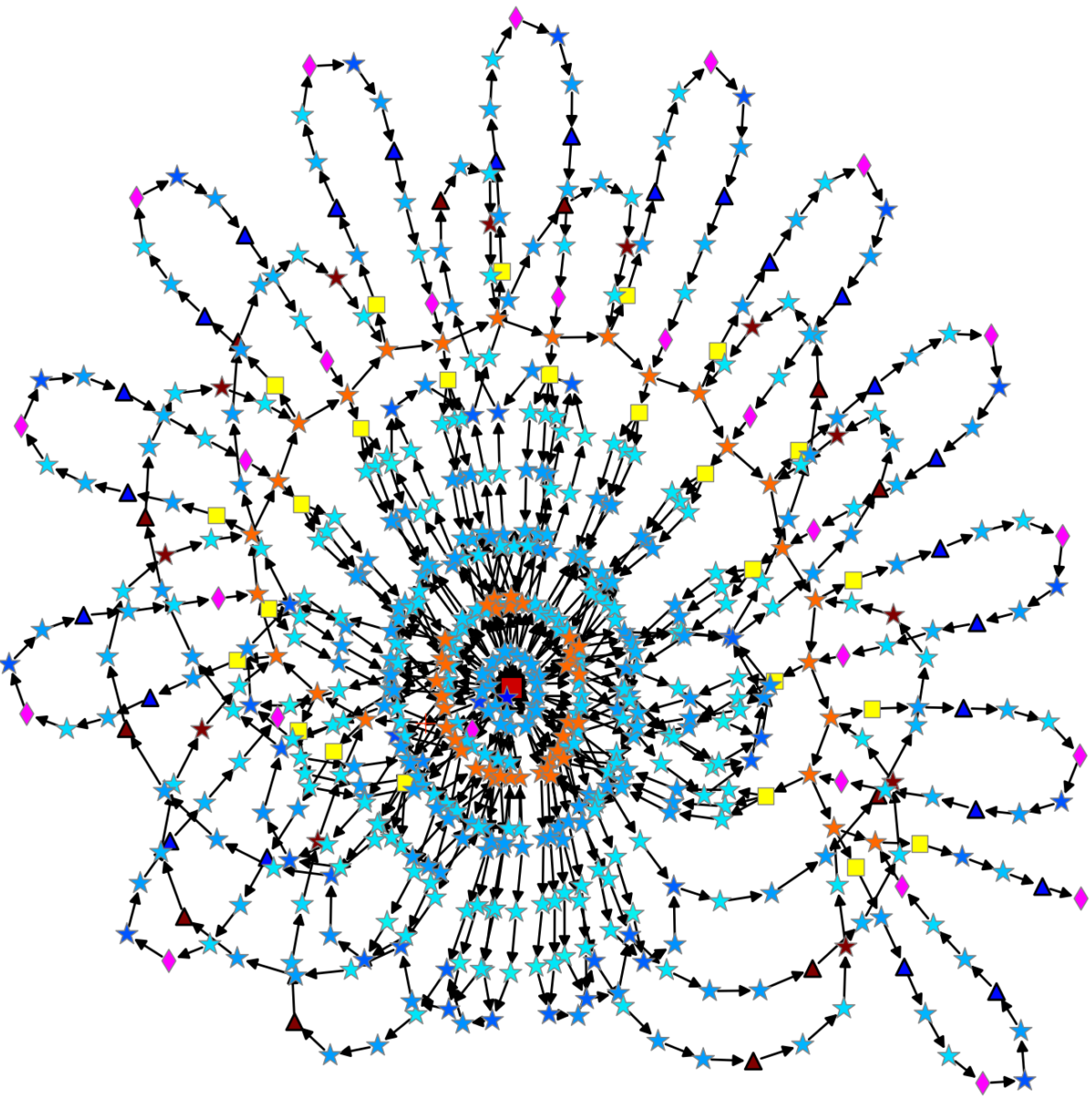}
    \caption{Vit-16-b}
  \end{subfigure}
  \begin{subfigure}[h!]{.3\textwidth}
    \centering\includegraphics[width=\textwidth]{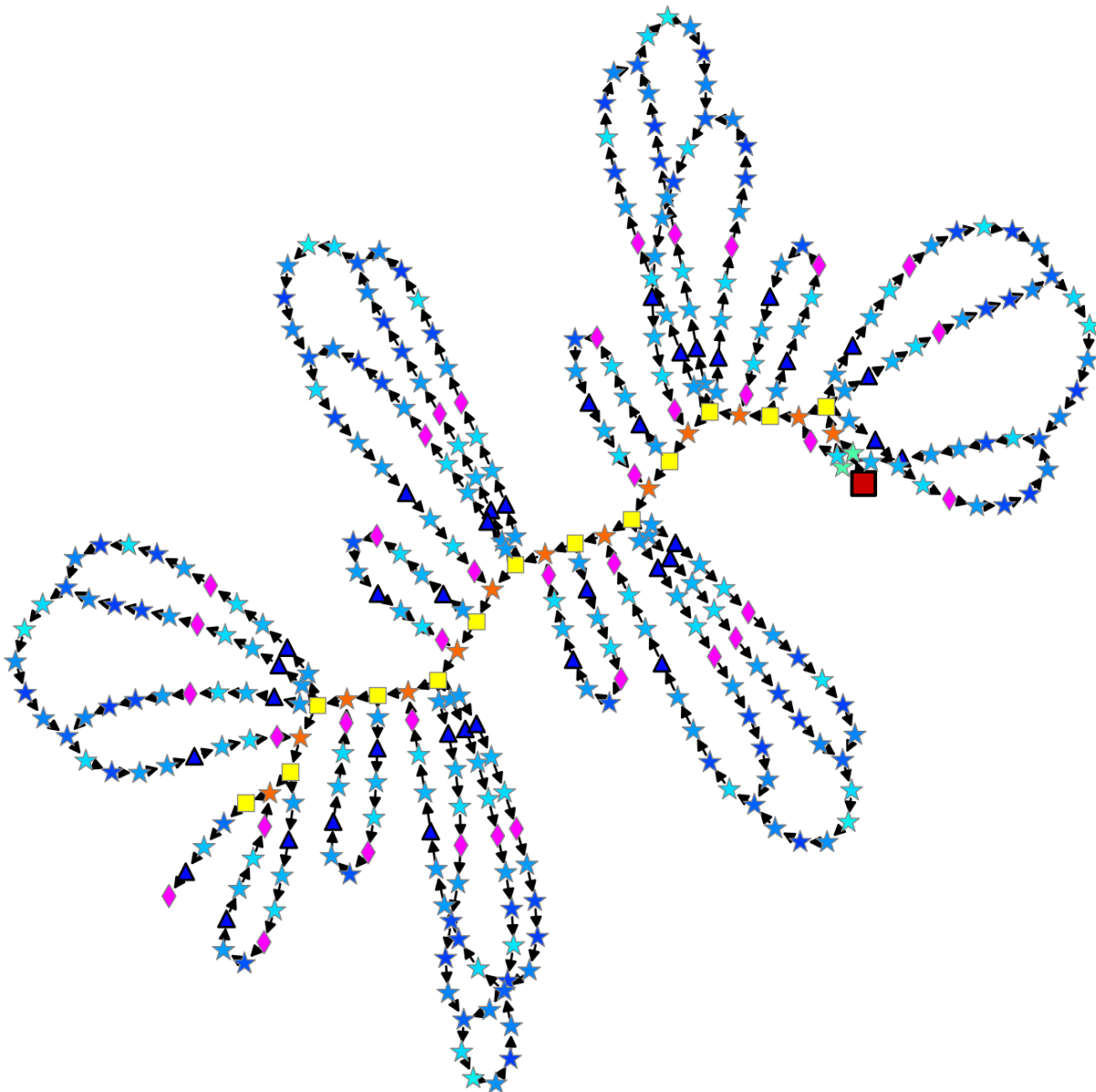}
    \caption{BERT-base}
  \end{subfigure}
  \begin{subfigure}[h!]{.3\textwidth}
    \centering\includegraphics[width=\textwidth]{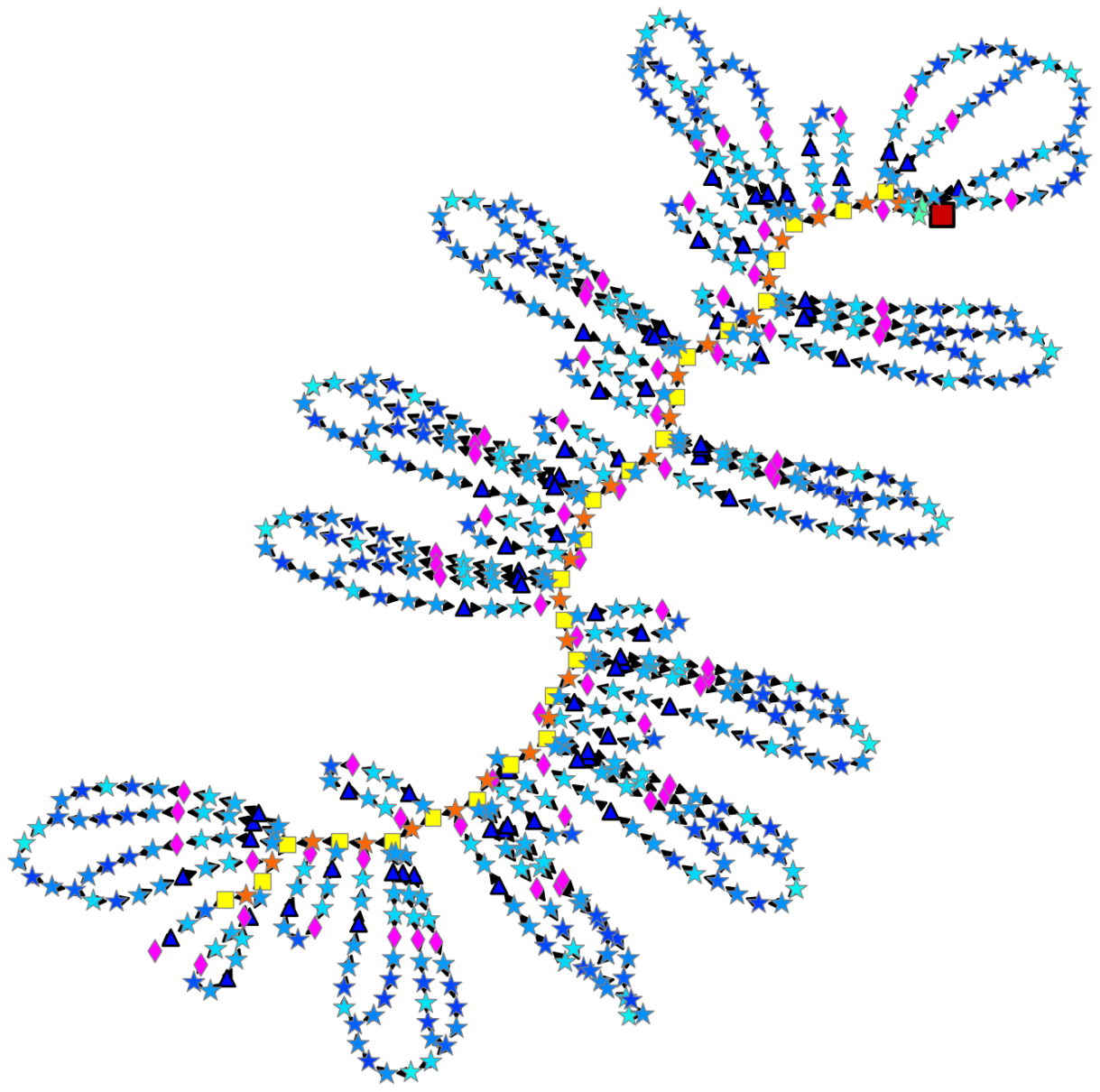}
    \caption{RoBERT-small}
  \end{subfigure}
  \begin{subfigure}[h!]{.3\textwidth}
    \centering\includegraphics[width=\textwidth]{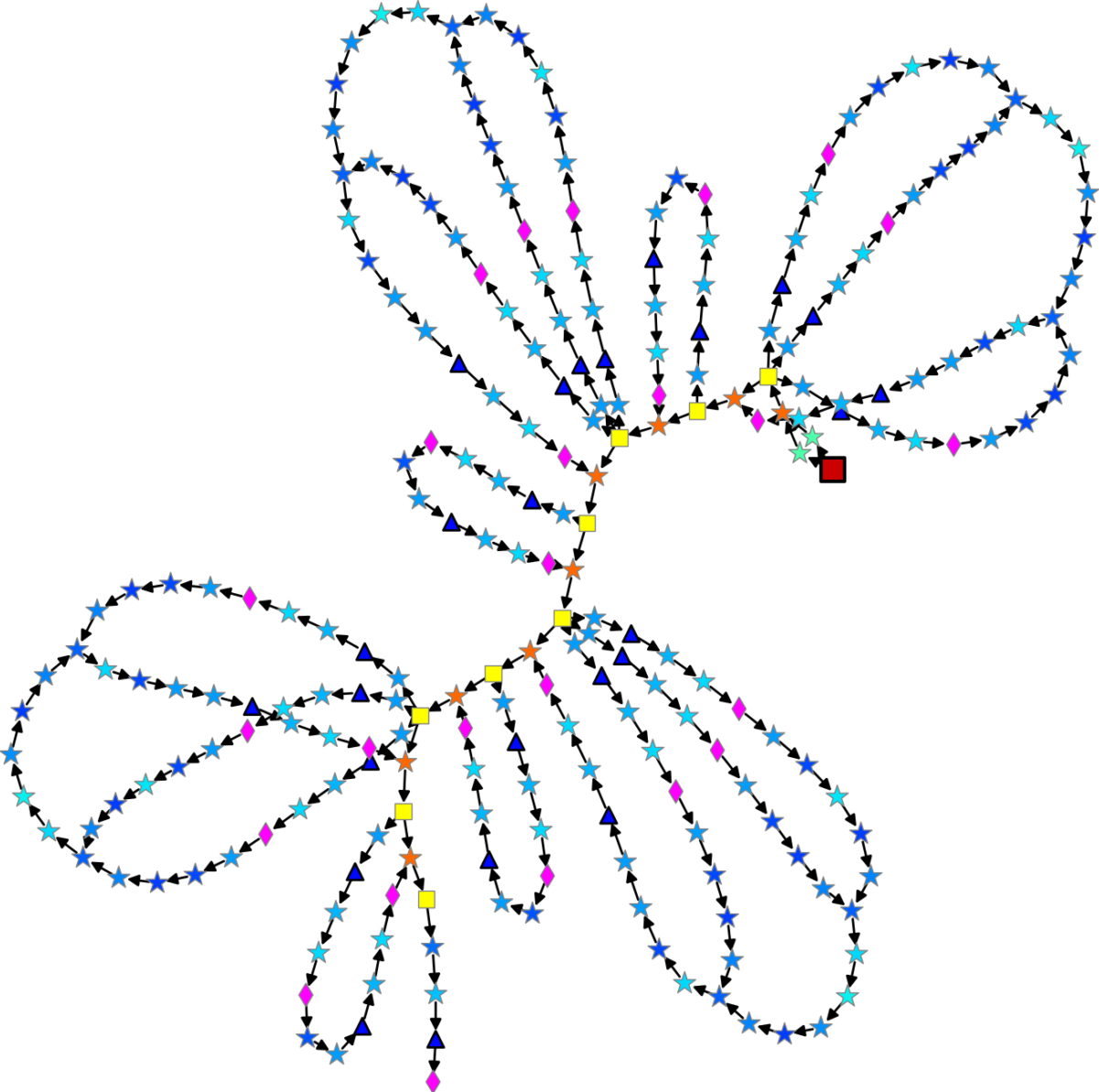}
    \caption{BERT-mini}
  \end{subfigure}
  \begin{subfigure}[h!]{.3\textwidth}
    \centering\includegraphics[width=\textwidth]{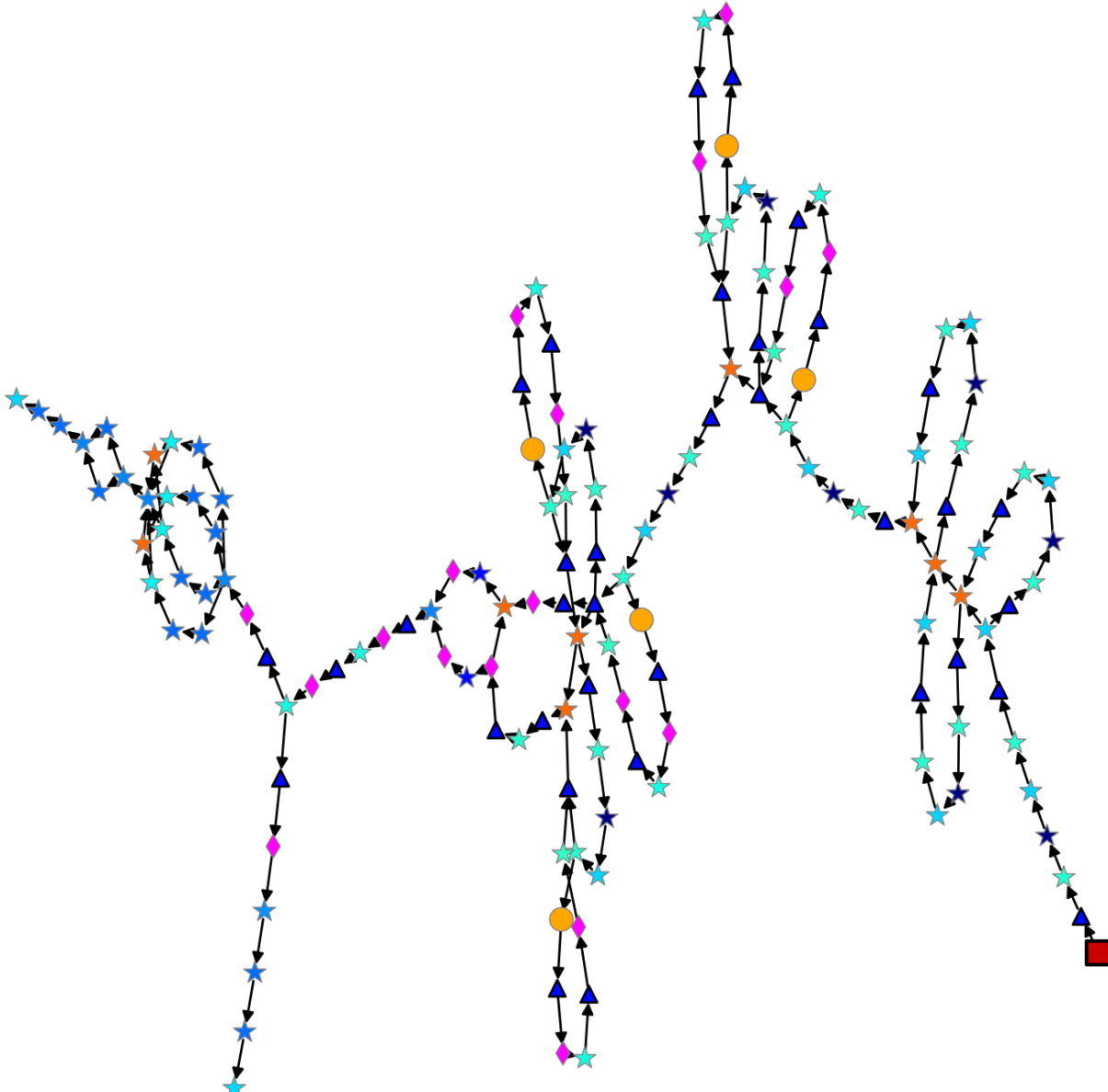}
    \caption{Fasterrcnn-MobileNet-Large-FPN}
  \end{subfigure}  
  \begin{subfigure}[h!]{1\textwidth}
    \centering\includegraphics[width=.6\textwidth]{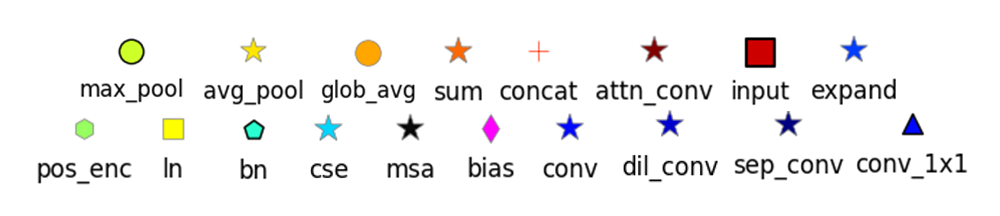}
  \end{subfigure}   
  
  \caption{Graphs generated for the architectures listed in Table \ref{tbl:sample-results}
      }
      \label{fig:tab5-graph}

\end{figure*}

\begin{figure*}
  \centering
  \begin{subfigure}[h!]{.4\textwidth}
    \centering\includegraphics[width=\textwidth]{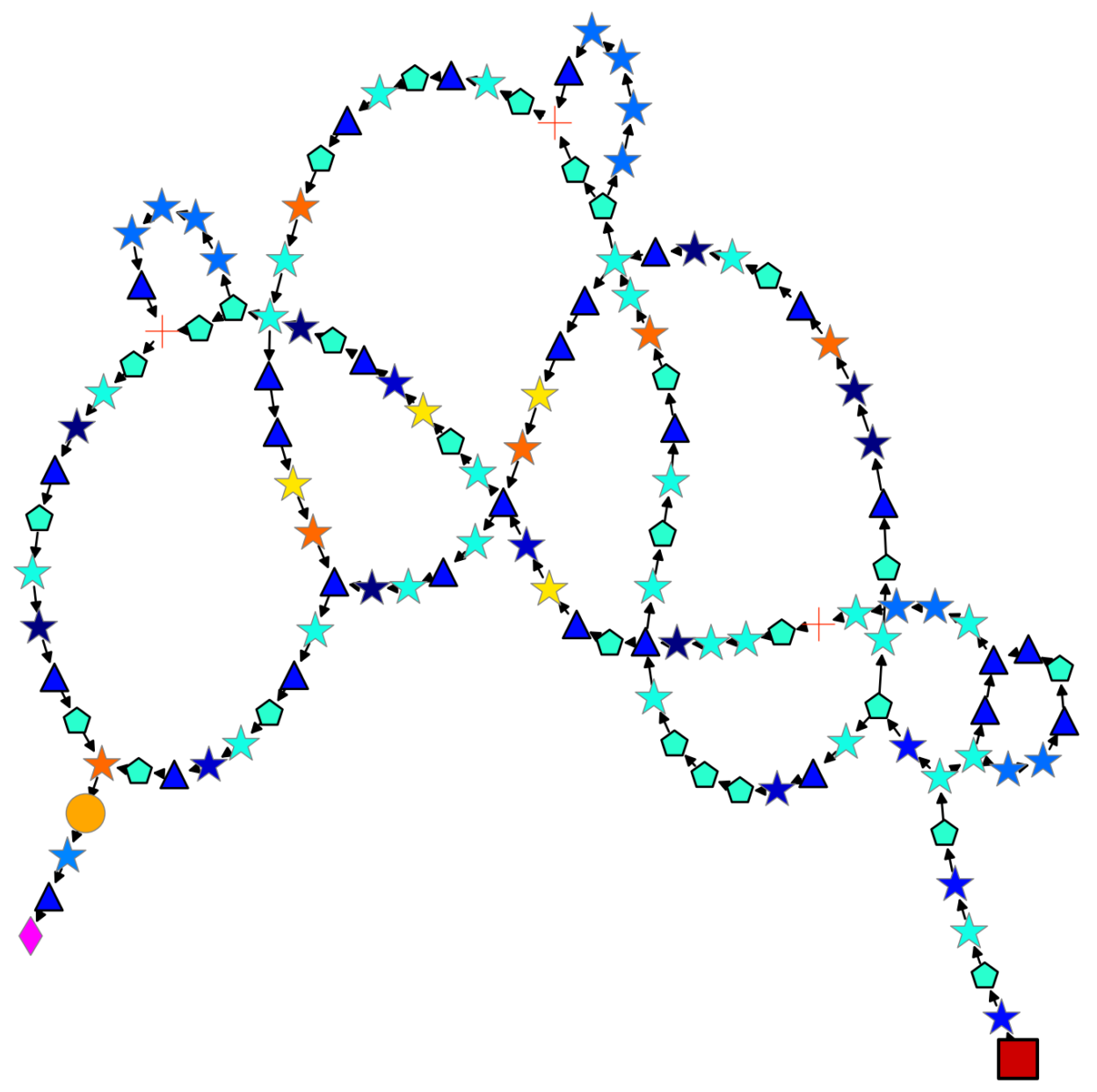}
    \caption{Architecture with layers: Conv2d, BatchNorm2d, ReLU, Dil\_conv2d, Sep\_conv2d, AvgPool2d, AdaptiveAvgPool2d, Linear}
  \end{subfigure}
  \\
  \begin{subfigure}[h!]{.4\textwidth}
    \centering\includegraphics[width=\textwidth]{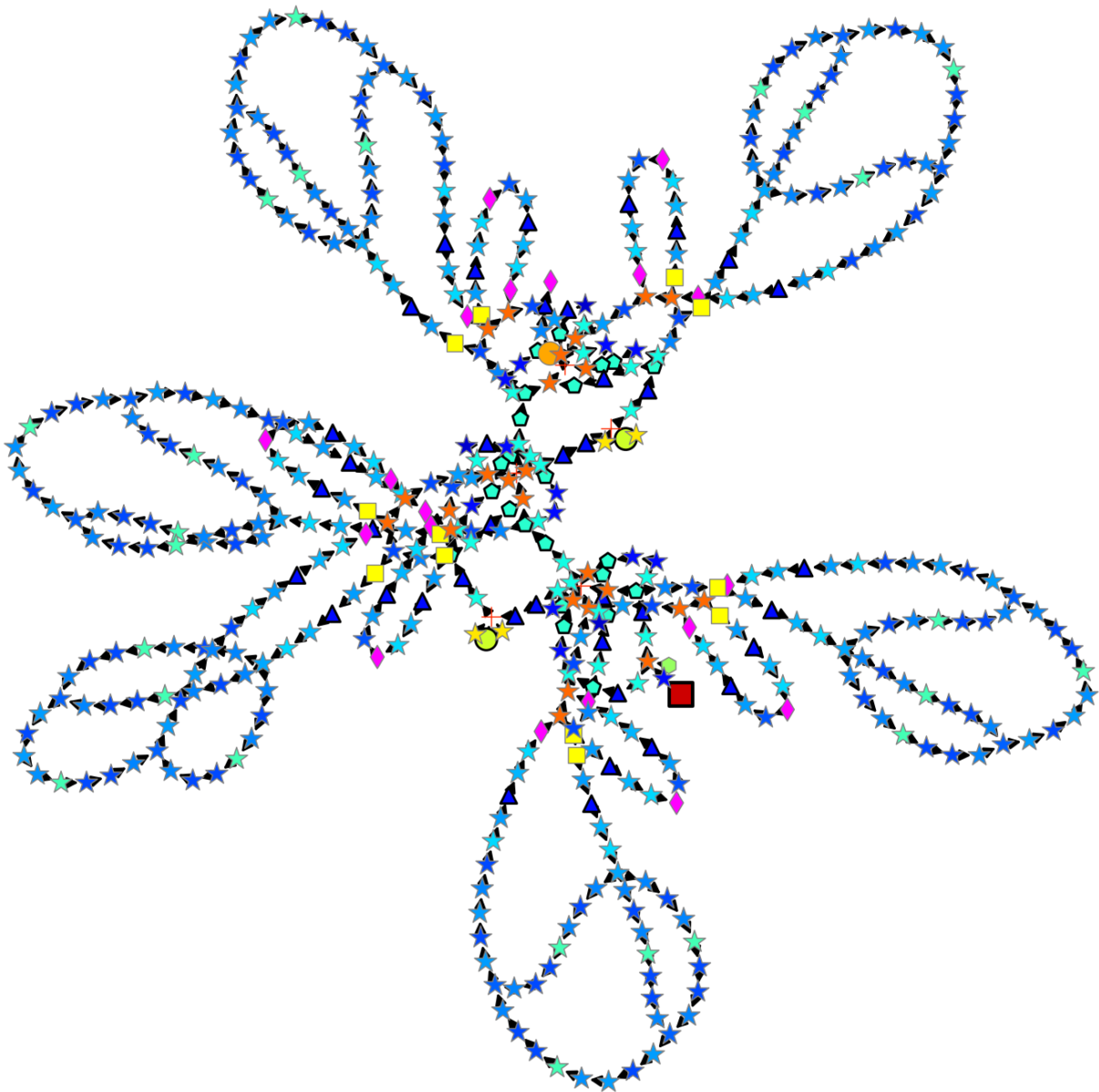}
    \caption{Architecture with layers: Conv2d, PosEnc, ReLU, BatchNorm2d, Linear, Dropout, LayerNorm, GELU, Dil\_conv2d, Zero, MaxPool2d, AvgPool2d, AdaptiveAvgPool2d}
  \end{subfigure}

    \begin{subfigure}[h!]{1\textwidth}
    \centering\includegraphics[width=.6\textwidth]{legend.PNG}
  \end{subfigure}   
  
  \caption{Graphs generated for the architectures listed in Tables \ref{tbl:datasets-AutoNet} and \ref{tbl:datasets-AQA}.}
  \label{fig:tab-autonet-graph}

\end{figure*}

In Figures \ref{fig:tab5-graph} and \ref{fig:tab-autonet-graph}, the visualization of all graphs generated for the neural architectures listed in Tables \ref{tbl:sample-results}, \ref{tbl:datasets-AutoNet}, and \ref{tbl:datasets-AQA} are illustrated.

\subsection{Dataset Quality Analysis}
We provide dataset quality analysis based on four criteria: reliability and completeness, label/feature noise, feature representation, and minimizing skew \citep{data_quality}.

\subsubsection{Reliability and Completeness}
The reliability of data refers to how trustable the data is, whether it has duplicated values and if it covers both positive and negative samples. As for dataset completeness, it refers to how much of the relevant information is included in the dataset for dealing with the desired problem.

In our TVHF dataset, we have collected models and their relevant descriptions as related bi-modal data types for the ArchBERT model to learn neural architectures along with their corresponding natural language descriptions. We considered the reliability and completeness of our dataset by collecting various models with different architectures designed for different tasks such as image and text classification, object detection, text summarization, etc. Also, the descriptions that have been assigned to each model were collected through blog posts, articles, papers, and documentations containing both high/low-level information related to that specific model. Due to the limited number of human-designed models, to make our dataset large enough for training purposes, we used each architecture more than once, and each time we assigned a different unique description to it to avoid having duplicate architecture-description pairs in our dataset. Moreover, we generated negative samples by assigning irrelevant descriptions to the architectures, so that the model could learn both similarities and dissimilarities.

As discussed in Section \ref{sec:datasets}, some of the descriptions in TVHF dataset did not include relevant technical information to the corresponding models. We manually reviewed the descriptions and removed such samples. We will further enhance the descriptions associated with each model within the release of the next version of our dataset.

\subsubsection{Label/Feature Noise}
Label noise refers to an imperfect annotation of data that confounds the assessment of model performance when training machine learning models. Feature noise can be defined as the noise got into the dataset through various factors such as incorrect collection by humans or instruments. Inconsistencies in data formats, missing values, and outliers are examples of noise created by this process.

If noise in a dataset is defined as a wrong description for a model, our dataset is a noise-free dataset because we annotated the samples manually.

Since the description of building blocks in the AutoNet models are converted to textual descriptions and question samples automatically, all the generated samples are relevant and noise-free.

For our ACD dataset, we manually hard-labeled the models based on their similarity with each another. Therefore, there is no missed or wrongly labeled example in the entire dataset.

\subsubsection{Feature Representation}
Mapping data to useful features while presenting them to the model is defined as feature representation. In this case, we consider how data is presented to the model and whether the numeric values need to be normalized.

To show our data to the ArchBERT model, we have been consistent in the following way. For architectures, based on their computational graphs, we extracted nodes, shapes, and edges, which the major and sufficient items to represent an architecture in our work. We then normalized these items and passed them to the model. As for descriptions, we represented each textual description with tokens, normalized them, and used them as inputs to the model. 

\subsubsection{Minimizing Skew}
One of the reasons that may cause getting different results for computed metrics at training vs. validation stages is training/validation skew. It usually happens when different features are presented to the model in training and validation stages.

We have collected our data and presented them to the model in the way that both training and validation stages receive the exact same set of features coming from the same distribution. This guarantees that our data is not skewed towards training or validation stages.

\begin{table*}
\small
\caption{Positive data samples from the TVHF dataset (TV: TorchVision, HF: HuggingFace).}
\centering
\newenvironment{mytab}[1][*{80}{c}]
\resizebox{\textwidth}{}{%
\begin{tabular}{c|l|c}
\hlineB{3}
\textbf{Architecture} &
  \multicolumn{1}{c|}{\textbf{Description}} &
  \textbf{Source} \\ \hline
vit\_b\_16 &
  adopted from BERT &
  \begin{tabular}[c]{@{}c@{}}TV\end{tabular} \\ \hline
segmentation.deeplabv3\_resnet101 &
  \begin{tabular}[c]{@{}l@{}}Improved version of DeepLab v2, with optimi-\\zation of ASPP layer hyper parameters and\\without a Dense CRF layer, for faster operation.\end{tabular} &
  \begin{tabular}[c]{@{}c@{}}TV\end{tabular} \\ \hline
resnet101 &
  \begin{tabular}[c]{@{}l@{}}Residual Networks, or ResNets, learn \\residual functions with reference to the \\layer inputs , instead of learning unreferenced \\functions.\end{tabular} &
  \begin{tabular}[c]{@{}c@{}}TV\end{tabular} \\ \hline
densenet121 &
  \begin{tabular}[c]{@{}l@{}}A DenseNet is a type of convolutional \\neural network that utilises dense connections \\between layers, through Dense Blocks, \\where we connect all layers (with matching \\ feature-map sizes) directly with each other\end{tabular} &
  \begin{tabular}[c]{@{}c@{}}TV\end{tabular} \\ \hline
resnext50\_32x4d &
  \begin{tabular}[c]{@{}l@{}}ResNeXt is a homogeneous neural network \\which reduces the number of hyper parameters \\required by conventional ResNet.\end{tabular} &
  \begin{tabular}[c]{@{}c@{}}TV\end{tabular} \\ \hline
detection.keypointrcnn\_resnet50\_fpn &
  \begin{tabular}[c]{@{}l@{}}12 Million Parameters, 2 Billion FLOPs and \\File Size is 47.08  MB.\end{tabular} &
  \multicolumn{1}{c}{\begin{tabular}[c]{@{}c@{}}TV\end{tabular}} \\ \hline
DemangeJeremy/4-sentiments-with-flaubert &
  \begin{tabular}[c]{@{}l@{}}This model is a fine-tuned version of \\google/fnet-base on the GLUE WNLI dataset\end{tabular} &
  HF \\ \hline
ctoraman/RoBERTa-TR-medium-char &
  \begin{tabular}[c]{@{}l@{}}Model architecture is similar to bert-medium \\ (8 layers, 8 heads, and 512 hidden size)\end{tabular} &
  HF \\ \hline
google/t5-efficient-base-dm1000 &
  \begin{tabular}[c]{@{}l@{}}T5-Efficient-BASE-DM1000 is a variation of \\Google's original T5 following the T5 model \\architecture.\end{tabular} &
  HF \\ \hline
microsoft/unihanlm-base &
  \begin{tabular}[c]{@{}l@{}}a self-supervised Chinese-Japanese pre-trained \\masked language model (MLM) with a novel \\two-stage coarse-to-fine training approach.\end{tabular} &
  HF \\ \hline
facebook/wmt21-dense-24-wide-en-x &
  \begin{tabular}[c]{@{}l@{}}WMT 21 En-X is a 4.7B multilingual \\encoder-decoder (seq-to-seq) model trained \\for one-to-many multilingual translation.\end{tabular} &
  HF \\ \hlineB{3}
\end{tabular}%
}
\label{tbl:datasets-TVHF}
\end{table*}

\begin{table*}
\caption{Positive and negative data samples from TVHF-ACD validation set (TV: TorchVision, HF: HuggingFace, 0: dissimilar, 1: similar).}
\centering
\newenvironment{mytab}[1][*{80}{c}]
\resizebox{\textwidth}{}{%

\begin{tabular}{l|l|c|c}
\hlineB{3}
\multicolumn{1}{c|}{\textbf{Architecture 1}}                         & \multicolumn{1}{c|}{\textbf{Architecture 2}}                        & \multicolumn{1}{c|}{\textbf{Label}} & \multicolumn{1}{c}{\textbf{Source}} \\ \hline
vgg11                           & vgg19\_bn                      & 1                        & TV           \\ \hline
mnasnet0\_5                     & mnasnet0\_75                   & 1                        & TV           \\ \hline
inception\_v3                   & efficientnet\_b3               & 0                     & TV           \\ \hline
efficientnet\_b1                & regnet\_x\_800mf               & 0                     & TV           \\ \hline
google/t5-efficient-large-kv128 & google/t5-efficient-small-kv16 & 1                        & HF                       \\ \hline
jweb/japanese-soseki-gpt2-1b    & tartuNLP/gpt-4-est-large       & 1                        & HF                       \\ \hline
hakurei/gpt-j-random-tinier     & minimaxir/magic-the-gathering  & 0                     & HF                       \\ \hline
mwesner/bart-mlm                & tartuNLP/gpt-4-est-base        & 0                     &             HF           \\ \hlineB{3}
\end{tabular}%
}
\label{tbl:datasets-CD}
\end{table*}

\begin{table*}
\caption{Positive and negative data samples from AutoNet (Architecture: list of unique layers, 0: negative sample, 1: positive sample).}
\centering
\resizebox{\textwidth}{!}{%
\begin{tabular}{c|l|c}
\hlineB{3}
Architecture &
  \multicolumn{1}{c|}{Description} & {Label}  
  \\ \hline
\multirow{5}{*}{\begin{tabular}[c]{@{}c@{}}'Conv2d', \\ 'PosEnc', \\ 'ReLU', \\ 'BatchNorm2d', \\ 'Linear', \\ 'Dropout', \\ 'LayerNorm', \\ 'GELU', \\ 'Dil\_conv2d', \\ 'Zero', \\ 'MaxPool2d', \\ 'AvgPool2d', \\ 'AdaptiveAvgPool2d'\end{tabular}} &
  \begin{tabular}[c]{@{}l@{}}This architecture contains 2d max pooling layer which is a pooling \\ operation that calculates the maximum value, and Gaussian Error Linear \\ Units (gelu) activation function which is a smoother version of RELU. \\ It also has 2D Adaptive Average pooling layer.  \end{tabular} & 1 \\ \cline{2-3} 
 &
  \begin{tabular}[c]{@{}l@{}}This neural network has Layer normalization over input across the features \\ instead of batch dimension, and linear module which applies a linear \\ transformation to the incoming data. It also contains Dropout layer that is \\ used to drastically reduce the chance of overfitting during training.\end{tabular} & 1\\ \cline{2-3} 
 &
  \begin{tabular}[c]{@{}l@{}}This classification neural network includes 2D average pooling layer used \\ for calculating the average for each patch of the feature map and has \\ about 1.18 Million parameters. In Totall, this neural network architecture has \\ 432 layers, and, it has 95 Conv2d, 1 PosEnc, 80 ReLU, 79 BatchNorm2d, \\ 62 Linear, 46 Dropout, 30 LayerNorm, 15 GELU, 15 Dil\_conv2d, 4 Zero, \\ 2 MaxPool2d, 2 AvgPool2d, and 1 AdaptiveAvgPool2d layer.\end{tabular} & 1\\ \cline{2-3} 
 &
  \begin{tabular}[c]{@{}l@{}}This neural architecture has 2D frozen batch normalization module in which \\ the batch statistics and the affine parameters are fixed, and Anchor \\ Generator module which is a standard for 2D anchor-based detectors. \\ Additionally, this architecture contains stochastic depth layer which aims to \\ shrink the depth of a network during training.\end{tabular}  & 0\\ \cline{2-3} 
 &
  \begin{tabular}[c]{@{}l@{}}This classifier includes 2D transposed convolution layer that applies convolution \\ with a fractional stride.\end{tabular}   & 0 \\ \hline
\multirow{4}{*}{\begin{tabular}[c]{@{}c@{}}'Conv2d', \\ 'Hardswish', \\ 'GeLU', \\ 'AvgPool2d', \\ 'Sep\_conv2d', '\\ AdaptiveAvgPool2d', \\ 'Dropout' \end{tabular}} &
  \begin{tabular}[c]{@{}l@{}}This classification neural architecture has Separable Convolution which \\ divides a single convolution into two or more convolutions to reduce the \\ number of parameters while producing the same output, and Hard Swish \\ activation function that replaces the computationally expensive sigmoid \\ with a piecewise linear analogue. This classifier also includes 2D average \\ pooling layer used for calculating the average for each patch of the feature \\ map.\end{tabular} & 1 \\ \cline{2-3} 
 &
  \begin{tabular}[c]{@{}l@{}}This network includes Dropout layer that is used to drastically reduce \\ the chance of overfitting during training, and 2D Adaptive Average pooling \\ layer. This neural architecture has about 0.38 Million parameters.\end{tabular} & 1 \\ \cline{2-3} 
 &
  \begin{tabular}[c]{@{}l@{}}This classification architecture includes generalized rcnn transform \\ module which performs input transformation before feeding the data \\ to a GeneralizedRCNN model, and Quantize stub module that is a \\ place holder for quantize operation. Another part of this neural network \\ is ReLU6 activation function which is a modification of the rectified linear \\ unit (relu) where the activation is limited to a maximum size of 6.\end{tabular} & 0 \\ \cline{2-3} 
 &
  \begin{tabular}[c]{@{}l@{}}This architecture contains Layer normalization over input across the features \\ instead of batch dimension, and dequantization module which converts tensors \\ from quantized to floating point.\end{tabular} & 0\\ \hlineB{3}
\end{tabular}%
}
\label{tbl:datasets-AutoNet}
\end{table*}

\begin{table*}
\caption{Positive and negative data samples for BACD task (TV: TorchVision, HF: HuggingFace, 0: dissimilar, 1: similar).}
\centering
\resizebox{\textwidth}{!}{%
\begin{tabular}{l|l|l|c|c}
\hlineB{3}
\multicolumn{1}{c|}{\textbf{Architecture 1}} &
  \multicolumn{1}{c|}{\textbf{Architecture 2}} &
  \multicolumn{1}{c|}{\textbf{Supporting text}} &
  \textbf{Label} &
  \textbf{Source} \\ \hline
resnet18 &
  segmentation.fcn\_resnet101 &
  \begin{tabular}[c]{@{}l@{}}A model containing \\ residual connection\end{tabular} &
  1 &
  TV \\ \hline
mnasnet0\_5 &
  vgg19 &
  \begin{tabular}[c]{@{}l@{}}An architecture for \\ image classification\end{tabular} &
  1&
  TV \\ \hline
wide\_resnet101\_2 &
  segmentation.deeplabv3\_resnet50 &
  \begin{tabular}[c]{@{}l@{}}An architecture for \\ image classification\end{tabular} &
  \begin{tabular}[c]{@{}c@{}}0\end{tabular} &
  TV \\ \hline
resnet34 &
  alexnet &
  \begin{tabular}[c]{@{}l@{}}A model containing \\ residual connection\end{tabular} &
  \begin{tabular}[c]{@{}c@{}}0\end{tabular} &
  TV \\ \hline
\begin{tabular}[c]{@{}l@{}}ctoraman/\\ RoBERTa-TR-medium-char\end{tabular} &
  \begin{tabular}[c]{@{}l@{}}ctoraman/\\ RoBERTa-TR-medium-wp-66k\end{tabular} &
  \begin{tabular}[c]{@{}l@{}}Model architecture is \\ similar to bert-medium\end{tabular} &
  1 &
  HF \\ \hline
\begin{tabular}[c]{@{}l@{}}dbmdz/\\ electra-base-turkish-cased-discriminator\end{tabular} &
  \begin{tabular}[c]{@{}l@{}}skplanet/\\ dialog-koelectra-small-generator\end{tabular} &
  \begin{tabular}[c]{@{}l@{}}containing ELECTRA \\ for self-supervised \\ language representation \\ learning\end{tabular} &
  1 &
  HF \\ \hline
\begin{tabular}[c]{@{}l@{}}rmihaylov/\\ pegasus-base-cnn-dailymail-bg\end{tabular} &
  TristanBehrens/js-fakes-4bars &
  \begin{tabular}[c]{@{}l@{}}A model for \\ summarization\end{tabular} &
  \begin{tabular}[c]{@{}c@{}}0\end{tabular} &
  HF \\ \hline
\begin{tabular}[c]{@{}l@{}}facebook/\\ m2m100-12B-avg-10-ckpt\end{tabular} &
  google/t5-11b-ssm-nqo &
  \begin{tabular}[c]{@{}l@{}}A pre-trained model \\ for Question Answering\end{tabular} &
  \begin{tabular}[c]{@{}c@{}}0\end{tabular} &
  HF \\ \hlineB{3}
\end{tabular}%
}
\label{tbl:datasets-BCD}
\end{table*}

\begin{table*}
\caption{Data samples from AutoNet-AQA (Architecture: list of unique layers).}
\centering
\resizebox{\textwidth}{!}{%
\begin{tabular}{c|l|l}
\hlineB{3}
Architecture &
  \multicolumn{1}{c|}{Question} &
  \multicolumn{1}{c}{Ground Truth Answer} \\ \hline
\multirow{5}{*}{\begin{tabular}[c]{@{}c@{}}Conv2d, \\ BatchNorm2d, \\ ReLU, \\ Dil\_conv2d, \\ Sep\_conv2d, \\ AvgPool2d, \\ AdaptiveAvgPool2d, \\ Linear\end{tabular}} &
  \begin{tabular}[c]{@{}l@{}}what type of pooling module has been used in \\ this neural architecture?\end{tabular} &
  \begin{tabular}[c]{@{}l@{}}AvgPool2d,\\ AdaptiveAvgPool2d\end{tabular} \\ \cline{2-3} 
 &
  \begin{tabular}[c]{@{}l@{}}what 2d average pooling layer performs in \\ this neural network?\end{tabular} &
  \begin{tabular}[c]{@{}l@{}}calculating the average for each \\ patch of the feature map\end{tabular} \\ \cline{2-3} 
 &
  \begin{tabular}[c]{@{}l@{}}what 2d Dilated Convolution module does in \\ this network?\end{tabular} &
  \begin{tabular}[c]{@{}l@{}}creating a wider kernel by inserting \\ spaces between the kernel elements\end{tabular} \\ \cline{2-3} 
 &
  \begin{tabular}[c]{@{}l@{}}what 2d max pool kernel size has been used in \\ this network?\end{tabular} &
  \begin{tabular}[c]{@{}l@{}}This model does not include \\ MaxPool2d\end{tabular} \\ \cline{2-3} 
 &
  \begin{tabular}[c]{@{}l@{}}in general what kernel size are used in this \\ neural network model?\end{tabular} &
  5*5,1*1,3*3 \\ \hline
\multirow{5}{*}{\begin{tabular}[c]{@{}c@{}}'Conv2d', \\ 'GELU', \\ 'MaxPool2d', \\ 'LayerNorm', \\ 'Linear', \\ 'Hardswish'\\ 'Dil\_conv2d', \\ 'LayerNorm'\end{tabular}} &
  \begin{tabular}[c]{@{}l@{}}what 2d max pooling module calculates in \\ this neural network?\end{tabular} &
  \begin{tabular}[c]{@{}l@{}}calculating the maximum value \\ for each patch of the feature map\end{tabular} \\ \cline{2-3} 
 &
  \begin{tabular}[c]{@{}l@{}}what type of normalization layer is used after \\ convolution  in this neural network architecture?\end{tabular} &
  LayerNorm \\ \cline{2-3} 
 &
  \begin{tabular}[c]{@{}l@{}}what type of activation layer has been used in \\ this neural network model?\end{tabular} &
  GELU, Hardswish \\ \cline{2-3} 
 &
  what hard sigmoid module performs in this model? &
  \begin{tabular}[c]{@{}l@{}}This model does not include \\ Hardsigmoid\end{tabular} \\ \cline{2-3} 
 &
  \begin{tabular}[c]{@{}l@{}}overall what kind of layers are included in this \\ neural network architecture?\end{tabular} &
  \begin{tabular}[c]{@{}l@{}}'Conv2d', 'GELU', 'MaxPool2d', \\ 'LayerNorm', 'Linear', 'Hardswish' \\ 'Dil\_conv2d', 'LayerNorm'\end{tabular} \\ \hlineB{3}
\end{tabular}%
}
\label{tbl:datasets-AQA}
\end{table*}

\end{document}